\documentclass[letterpaper, 10 pt, conference]{ieeeconf}  %

\IEEEoverridecommandlockouts                              %

\def\outputstyle{arxiv}

\def\elsevier{elsevier}

\usepackage{epsfig} %
\usepackage{amsmath} %
\usepackage{graphicx}
\ifx\outputstyle\elsevier
  \usepackage[labelfont=bf,justification=justified]{caption}
  \usepackage{subfig}
\else
  \usepackage[caption=false,font=footnotesize]{subfig}
\fi
\usepackage{listings}
\ifx\outputstyle\elsevier
\else
  \makeatletter
  \def\lst@makecaption{%
    \def\@captype{table}%
    \@makecaption
  }
  \makeatother
\fi
\usepackage{tikz}
\usepackage[simplified]{pgf-umlcd}
\usetikzlibrary{arrows.meta}
\usetikzlibrary{calc}
\usetikzlibrary{shadows}
\usepackage{xspace}
\usepackage[hidelinks]{hyperref}
\usepackage[nameinlink,capitalise]{cleveref}

\xspaceremoveexception{-}
\title{\LARGE \bf
Message Flow Analysis with Complex Causal \\ Links for Distributed ROS 2 Systems
}

\author{Christophe Bédard, Pierre-Yves Lajoie, Giovanni Beltrame, Michel Dagenais%
\thanks{The financial support of Ericsson, NSERC, Prompt, and Vanier Canada Graduate Scholarship is gratefully acknowledged.}%
\thanks{Department of Computer Engineering and Software Engineering, Polytechnique Montréal, Montreal, Quebec H3T 1J4, Canada,
        {\tt\scriptsize\{christophe.bedard, pierre-yves.lajoie, giovanni.beltrame,}
        {\tt\scriptsize michel.dagenais\}@polymtl.ca}}%
}

\IEEEtriggeratref{39}

\begin{document}

\newcommand{\rostwotracing}{\texttt{ros2\_tracing}\xspace}
\newcommand{\ROSone}{ROS~1\xspace}
\newcommand{\ROStwo}{ROS~2\xspace}
\newcommand{\tracetoolsanalysis}{\texttt{tracetools\_analysis}\xspace}
\newcommand{\performancetest}{performance\_test\xspace}

\newcommand{\etal}{et al.\xspace}
\newcommand{\note}[1]{\textcolor{blue}{#1}}

\newcommand{\vspacefiguretikz}{\vspace{-0.3cm}}
\newcommand{\vspacefigureimage}{}%
\newcommand{\vspacetablefootnotes}{}%
\newcommand{\vspacelisting}{}%

\ifx\outputstyle\elsevier
    \newcommand{\keywordsep}{\sep}
    \newcommand{\keywordend}{}
\else
    \newcommand{\keywordsep}{,\xspace}
    \newcommand{\keywordend}{.}
\fi
 
\maketitle
\thispagestyle{empty}
\pagestyle{empty}

\begin{abstract}
Distributed robotic systems rely heavily on the publish-subscribe communication paradigm and middleware frameworks that support it, such as the Robot Operating System (ROS), to efficiently implement modular computation graphs.
The \ROStwo executor, a high-level task scheduler which handles \ROStwo messages, is a performance bottleneck.
We extend \rostwotracing, a framework with instrumentation and tools for real-time tracing of \ROStwo, with the analysis and visualization of the flow of messages across distributed \ROStwo systems.
Our method detects one-to-many and many-to-many causal links between input and output messages, including indirect causal links through simple user-level annotations.
We validate our method on both synthetic and real robotic systems, and demonstrate its low runtime overhead.
Moreover, the underlying intermediate execution representation database can be further leveraged to extract additional metrics and high-level results.
This can provide valuable timing and scheduling information to further study and improve the \ROStwo executor as well as optimize any \ROStwo system.
The source code is available at: \href{https://github.com/christophebedard/ros2-message-flow-analysis}{github.com/christophebedard/ros2-message-flow-analysis}.
 \end{abstract}


\section{Introduction}\label{sec:introduction}

Modern robotic systems often leverage complex distributed processing: they use distributed perception~\cite{lajoieTowards2022}, motion planning~\cite{Dewangan2017}, and decision making~\cite{yan2013}.
They are built on software frameworks like \ROStwo~\cite{macenski2022ros2}, the successor to the Robot Operating System (ROS)~\cite{quigley2009ros}.
Such middleware frameworks greatly simplify the development of modular computation graphs.
However, high-level scheduling of tasks in \ROStwo (i.e., internal message handling, and subscription, service, or timer callback execution) brings a number of performance challenges.
Several methods and tools have been proposed to study the default \ROStwo executor and compare its performance with other proposed executor designs~\cite{lange2018callback,choi2021picas,staschulat2020rclc,staschulat2021budget}.
Furthermore, getting message latencies across a whole system is paramount to assessing the impact of these new approaches as well as evaluating the overall performance of a system, from one end to the other.
Tools have been introduced to study message latencies in real systems~\cite{nishimura2021raplet,li2022autoware_perf}.
However, these techniques are either not applicable to existing \ROStwo systems, since they require non-trivial code modifications, or result in significant runtime overhead.
Finally, existing techniques also cannot identify relationships between messages across possibly-distributed systems, which is necessary to extract the end-to-end latency.

Low-overhead tracing has been used as a way to extract execution information for performance analysis purposes without impacting or perturbing the system.
Furthermore, various techniques allow combining and correlating kernel and userspace events from multiple traces (i.e., from distributed systems).
In previous work, we proposed \rostwotracing~\cite{bedard2022ros2tracing}, a framework with low-overhead instrumentation and orchestration tools for tracing \ROStwo.
This tracing framework allows extracting \ROStwo execution information, and can be extended with additional instrumentation for more advanced use-cases or performance analysis goals, such as providing information about the overall performance of the message-passing system.

Moreover, critical path analysis has been used to model the interactions inside parallel and distributed systems.
For example, wait-related kernel events can be used to recursively compute wait dependencies across machines for requests spanning multiple hosts in a distributed system~\cite{giraldeau2015wait}.
This technique helps to understand and explain the actual process execution, and to identify the prime target for performance optimization (i.e., the bottleneck).
Likewise, in this paper, we present a message flow analysis method for \ROStwo distributed computation graphs.
It can be useful for both end users of \ROStwo, looking to analyze and optimize their system, and for developers, looking to do the same for the internals of \ROStwo, although these two target audiences are not necessarily distinct.

\begin{figure}[t]
\centerline{\includegraphics[width=\columnwidth,trim={0px 658px 2675px 35px},clip]{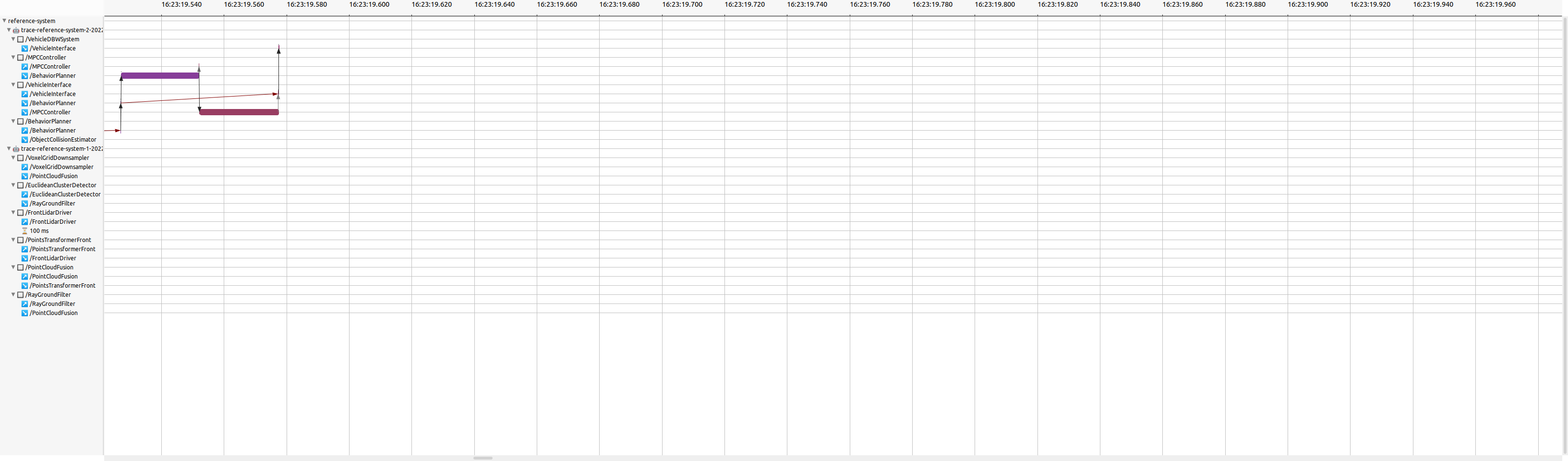}}
\vspacefigureimage
\caption{
Message flow visualization using our method.
}
\label{fig:introduction-method-example}
\end{figure}

\textbf{Contributions.}
As shown in \cref{fig:introduction-method-example}, our proposed technique can extract and visualize the paths of messages across distributed \ROStwo systems, providing information about the overall systems performance.
Building a message flow graph in a low-overhead way, without modifying the applications themselves as the existing methods do, requires more complex runtime execution data collection and analysis.
With this novel approach, we bring the following contributions:
\begin{itemize}
\item
    An intermediate execution representation database of trace data obtained from a distributed system, providing information on \ROStwo objects (i.e., nodes, publishers, subscriptions, and timers) and events (i.e., message publication and reception instances, and subscription and timer callback instances).
    This database can be further leveraged to derive additional metrics and high-level results.
\item Matching of published and received messages, compatible with distributed systems, without needing to modify the user code and without adding significant overhead.
\item Inference of one-to-one and one-to-many causal links between received messages and published messages, both automatically for direct links, and using simple user-level annotations for more complex indirect causal links.
\item Extraction and visualization of the flow of messages across distributed \ROStwo systems, as well as the state of the executor over time for each process.
\item Experiments and validation of our proposed method on both synthetic and real robotic systems, demonstrating that it can be used to study and optimize existing systems, and that it has a low runtime overhead.
\end{itemize}

This paper is structured as follows:
We first survey related work in \cref{sec:related-work}.
We then summarize relevant background information and describe our intermediate execution representation in \cref{sec:data-model}.
Thereafter, we present our analysis method in \cref{sec:analysis} and our executor visualization in \cref{sec:executor}.
Next, \cref{sec:experiments} presents experiments where we apply our method to two systems, and \cref{sec:overhead-evaluation} provides an evaluation of the runtime overhead.
Future work is outlined in \cref{sec:future-work}.
Finally, we conclude in \cref{sec:conclusion}.

\section{Related Work}\label{sec:related-work}

Previous work has identified open problems relating to the communications latency of \ROStwo~\cite{macenski2022ros2} and its executor.
Relevant methods were proposed to observe and study those problems.

\subsection{Communications}

First, the general performance of \ROStwo was evaluated by Maruyama \etal~\cite{maruyama2016exploring}, Gutiérrez \etal~\cite{gutierrez2018towards}, and Puck \etal~\cite{puck2020distributed}.
Other work focuses on more specific elements of the performance of \ROStwo, including its overhead with relation to the underlying middleware, DDS~\cite{pardo2003omg}.
Kronauer \etal~\cite{kronauer2021latency} evaluated the overhead of \ROStwo using profiling, and showed that it can lead to a 50\% latency overhead.
Some of this overhead can be attributed to the serialization and deserialization of complex message structures.
Jiang \etal~\cite{jiang2020message} proposed an adaptive serialization technique to improve communication performance by up to 93\%.
Wang \etal~\cite{wang2019tzc} proposed a single-host inter-process communications (IPC) layer for \ROSone~\cite{quigley2009ros} and \ROStwo which reduces the overhead of IPC for large messages.
Finally, Puck \etal~\cite{puck2021performance} noted in another performance evaluation of \ROStwo that the use of dynamic memory allocations, when fetching new messages from the underlying middleware, accounts for a significant portion of the internal message processing time.

Various tools have been proposed to study message latency in \ROStwo.
The \performancetest~\cite{performancetest} benchmarking tool allows measuring the latency between publishers and subscriptions directly, while~\cite{irobotros2performance} allows defining a custom message graph topology.
However, these benchmarking tools only evaluate the performance of a synthetic system or communication configuration.
To measure the performance of real systems, observability tools are needed.
Nishimura \etal~\cite{nishimura2021raplet} proposed RAPLET, which breaks down the latency between the publication of a message and the execution of the subscription callback function on the other end.
It tracks messages using a sequence number in the message structure itself.
Unfortunately, this sequence number field is not included in all messages.
Similarly, Witte and Tichy~\cite{witte2021inferred} presented a tool to track messages in \ROSone in order to interactively modify their content.
These techniques cannot be applied to existing systems, since they require the addition of a custom message header, and their runtime overhead is significant, which can affect the validity of their results~\cite{gregg2020systems}.

\subsection{Executor}\label{sec:related-work-executor}

Moreover, other previous work has identified and studied open problems with the ROS executor, which is a high-level task scheduler~\cite{ros2docsexecutor}.
It is responsible for fetching new messages from the underlying middleware and executing the corresponding subscription callbacks as well as timer callbacks, making the executor a clear performance bottleneck.
Furthermore, scheduling tasks on top of the OS scheduler itself is challenging.
Multiple methods model exchanges of ROS messages, from node to node, as event chains and pipelines in a directed acyclic graph (DAG).
Peeck \etal~\cite{peeck2021online} focused on online monitoring for reacting to latency violations in event chains.
Casini \etal~\cite{casini2019response} proposed a formal scheduling model and a response-time analysis for \ROStwo to bound worst-case response times.
Tang \etal~\cite{tang2020response} then proposed a more specific version that is, however, only valid for independent linear processing chains.
Blass \etal~\cite{blass2021automatic} built on the work by Casini \etal~\cite{casini2019response} and proposed an online automatic latency manager for \ROStwo.
Their work also helped illustrate how the higher-level scheduling of tasks in \ROStwo does not interact well with classic OS-level scheduling techniques.
Blass \etal~\cite{blass2021ros} further extended this work, and stressed how the \ROStwo executor differs from normal schedulers in the literature, since it inherently prioritizes in order: timers, subscriptions, service servers, and service clients.

To help tackle some of these challenges, new executor designs have been proposed for \ROStwo.
The callback-group-level executor~\cite{lange2018callback}, now available as an alternative in \ROStwo, allows having multiple distinct executor instances on multiple threads without interference.
This enables scheduling of the OS threads themselves, using different priorities depending on system requirements, instead of bundling all \ROStwo elements together, as the default executor does.
This results in lower latencies for higher-priority callback groups, as demonstrated by Yang and Azumi~\cite{yang2020exploring}.
Similarly, Choi \etal~\cite{choi2021picas} proposed a priority-driven chain-aware scheduler and showed that it helps lower end-to-end latencies as well.
Staschulat \etal~\cite{staschulat2020rclc,staschulat2021budget} proposed a budget-based executor for real-time operating systems.
To benchmark and compare executor designs, a reference system was proposed~\cite{rtwgreferencesystem}.
It is based on the computation graph of Autoware~\cite{kato2018autoware,kato2015open}, an autonomous driving system completely based on ROS.

\subsection{Tracing and Data Analysis}

To investigate performance issues, low-overhead tracing has been widely used for collecting execution information in a minimally-invasive way.
In particular, the LTTng tracer~\cite{desnoyers2006lttng}, which has a low runtime overhead~\cite{gebai2018survey}, was used by Lütkebohle~\cite{Luetkebohle2017} to investigate determinism and message timing issues in \ROSone.
They proposed~\cite{ros1tracetools} as a generic tracing tool for \ROSone.
As a follow-up to this for \ROStwo -- and to improve on it -- in previous work, we presented \rostwotracing~\cite{bedard2022ros2tracing}, a framework with low-overhead instrumentation and tracing tools for \ROStwo.
The proposed instrumentation can be used to extract simple metrics, such as publishing rate and subscription or timer callback duration, as demonstrated in~\cite{bedard2022ros2tracing}.
For other, more advanced performance analysis use-cases, the instrumentation can easily be extended.

To extract useful information from trace data, advanced trace analysis methods build models from the trace data.
One interesting technique is the critical path method, where the critical path is defined as the longest path in a DAG.
The critical path is therefore the overall program or end-to-end latency bottleneck; shortening it effectively reduces the total execution time.
Yang and Barton~\cite{yang1988critical} applied the method to compute the critical path of the execution of parallel and distributed programs.
Trace data from multiple hosts in a distributed system can be combined and synchronized for analysis as a whole~\cite{duda1987estimating,poirier2010accurate,jabbarifar2014liana}.
Giraldeau and Dagenais~\cite{giraldeau2015wait} used wait-related trace events from the kernel (e.g., scheduling, network, or interrupts) to recursively compute wait dependencies across machines.
While such wait-related operating system primitives are used in many applications, application-level information is required for more specialized analyses~\cite{gelle2021combining}.
For example, ROS-level information could be used to apply the critical path method to the computation graph of a ROS system.

Previous work has partially tackled this critical path analysis effort.
Santos \etal~\cite{santos2019static} used static code analysis to extract a model of the system architecture for \ROSone.
While it does not require executing the code, it only provides an overall view of the system architecture: it does not provide time-related information about the individual messages going through the system and the links between the messages.
Therefore, to consider possibly complex dynamic timing interactions, runtime execution information is required.
To this end, Li \etal~\cite{li2022autoware_perf} used \rostwotracing~\cite{bedard2022ros2tracing} and \tracetoolsanalysis~\cite{tracetoolsanalysis}, a simple trace data processing library, to provide an end-to-end latency breakdown.
However, links between input subscriptions and output publishers need to be manually provided by the user; they are not automatically detected.
\cite{ros1tracetools} was used and extended by~\cite{bedard2019messageflow} to visualize the flow of messages.
Unfortunately, \cite{bedard2019messageflow} has many limitations and uses simplistic assumptions which do not always hold.
For example, to track messages between nodes, it selects the first TCP packet that is queued after a message is sent by \ROSone.
It then matches that network packet when it is received on the other end, and selects the next message reception event.
This heuristic is simple and does not require adding additional fields to the messages themselves, but it is far from solid, since it could select packets from other applications.
Furthermore, it only considers direct causal links inside subscription callbacks, i.e., where the message being processed by the callback instance is linked to the message that is published during that callback instance.
However, many systems use custom message cache mechanisms that are independent from the \ROStwo API, which makes detecting and modeling those causal links far from trivial.
Finally, it does not support one-to-many or many-to-many causal links, i.e., where one or more input messages are linked to more than one output message, and also does not work with more than one machine.

\subsection{Summary}

In summary, numerous latency- and executor-related open problems exist in \ROStwo.
Building a model and graph of the path of messages across a \ROStwo system would provide useful information to further study or work on resolving those open problems.
Thus, we propose a low-overhead technique that can transparently and natively track messages while supporting complex application-dependent causal links between messages.

\section{Intermediate Execution Representation}\label{sec:data-model}

Before extracting the flow of \ROStwo messages across a distributed system, we first process the raw trace data to create a higher-level intermediate representation of the execution.
The underlying database can then be queried to build the actual message flow analysis; this process is greatly simplified by the intermediate representation.
The database can also be queried for other analysis purposes.

\subsection{\ROStwo Architecture}\label{sec:data-model-ros-architecture}

\begin{figure}[htbp]
\begin{center}
\tikzset{>={Latex[width=1.5mm,length=1.5mm]},
  baseTxt/.style = {scale=0.90, text centered, font=\sffamily},
  rosCoreTxt/.style = {baseTxt, font=\ttfamily},
  base/.style = {baseTxt, rectangle, draw=black, minimum width=2.0cm, minimum height=1cm},
  network/.style = {base, fill=red!20},
  dds/.style = {base, fill=yellow!20},
  rosCore/.style = {base, fill=violet!20},
  userPkg/.style = {base, fill=green!20},
  flowArrows/.style = {rounded corners=1mm}
}

\begin{tikzpicture}[align=center]
  \draw[network] (0,0.0)   rectangle (2.0,0.5)  node[baseTxt, pos=.5]    (network)      {network};
  \draw[network] (2.0,0.0) rectangle (3,0.5)    node[baseTxt, pos=.5]    (networkOther) {...};
  \draw[dds]     (0,0.5)   rectangle (2.0,1.0)  node[baseTxt, pos=.5]                   {DDS};
  \draw[dds]     (2.0,0.5) rectangle (3,1.0)    node[baseTxt, pos=.5]                   {...};
  \draw[rosCore] (0,1.0)   rectangle (3,1.5)    node[rosCoreTxt, pos=.5]                {rmw};
  \draw[rosCore] (0,1.5)   rectangle (3,2.0)    node[rosCoreTxt, pos=.5]                {rcl};
  \draw[rosCore] (0,2.0)   rectangle (1.5,2.5)  node[rosCoreTxt, pos=.5]                {rclcpp};
  \draw[rosCore] (1.5,2.0) rectangle (3,2.5)    node[rosCoreTxt, pos=.5]                {rclpy};
  \draw[userPkg] (0,2.5)   rectangle (3,3.0)    node[baseTxt, pos=.5]    (user)         {user code};
  
  \draw[->, flowArrows] ($(user.west)-(0.75,0)$)         |- ($(network.west)+(0.0,0)$);
  \draw[->, flowArrows] ($(networkOther.east)-(0.45,0)$) -| ($(user.east)+(0.75,0)$);
\end{tikzpicture}
 \end{center}
\vspacefiguretikz
\caption{
\ROStwo architecture and interactions between layers.
For example, a message is created by user code, goes through the layers of the \ROStwo architecture down to the network, and then goes back up to the user code on the other end (e.g., from one node to another node).
While DDS is the default middleware, other middlewares and transmission mechanisms can be used.
}
\label{fig:data-model-ros-2-architecture}
\end{figure}
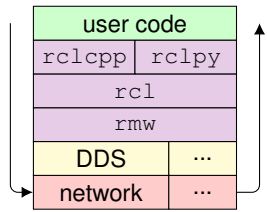

As shown in \cref{fig:data-model-ros-2-architecture}, \ROStwo contains multiple abstraction layers.
From top to bottom, i.e., from user-level to OS-level: \texttt{rclcpp} and \texttt{rclpy}, \texttt{rcl}, and \texttt{rmw}.
The client libraries, \texttt{rclcpp} and \texttt{rclpy}, offer the actual user-facing \ROStwo C++ and Python APIs, respectively.
They use a common underlying library, \texttt{rcl}; this architecture reduces duplicate code and thus makes adding new client libraries simpler.
Then, \texttt{rcl} calls \texttt{rmw}, the middleware interface.
This interface is implemented for each underlying middleware implementation, e.g., for each distinct DDS implementation.
This allows \ROStwo to use any message-passing middleware with any transmission mechanism, as long as it is done through this interface.
The flow of information, from the user code on one end to the user code on the other end, is illustrated in \cref{fig:data-model-ros-2-architecture}.

\subsection{Processing}\label{sec:data-model-processing}

Since the \ROStwo architecture contains multiple separate abstraction layers, the information collected by \rostwotracing~\cite{bedard2022ros2tracing} for analysis purposes is spread out over all layers.
This also provides internal information about \ROStwo.
For example, the duration of the message publication call can be broken down into \texttt{rclcpp}, \texttt{rcl}, and DDS time.
Hence, \rostwotracing instruments all layers in order to collect all relevant information.
Furthermore, to minimize the runtime impact, the \rostwotracing instrumentation is split into two distinct groups: initialization and runtime.
The initialization instrumentation points collect one-time information, in order to minimize the size of the data collected by the runtime instrumentation points, which are executed more often.
Therefore, we need to combine data from multiple instrumentation points in order to get the high-level information we need.
Moreover, \rostwotracing does not include instrumentation for the chosen DDS implementation; thus we instrumented it and combined all this information.

For example, when a new publisher object is created, 3 tracepoints are triggered: the first one is in \texttt{rcl}, the second one is in \texttt{rmw}, and the last one is in the DDS implementation.
By combining the execution information collected at different levels of the \ROStwo architecture by the 3 tracepoints into one publisher entry in the database, we can attribute the \texttt{rcl}-, \texttt{rmw}-, or DDS-level data, of the subsequent publication instance trace events, to the corresponding publisher.
To correlate and merge the information from multiple trace events, unique identifiers are required.
In most cases, \rostwotracing uses the values of the pointers to the underlying internal data structures, i.e., memory addresses.
To combine DDS-level information with the above \ROStwo information, we use the globally-unique identifier (GUID or GID) of the DDS data writer, which is the DDS-level object that actually sends messages from a \ROStwo publisher.
This GID is used internally in \ROStwo and is part of the \texttt{rmw} interface.
We have instrumented both eProsima Fast DDS~\cite{eprosimafastdds} and Eclipse Cyclone DDS~\cite{eclipsecyclone}.
As a result of the above design, either DDS implementation can be used without affecting the model.

Our intermediate representation needs to be valid when tracing multiple processes on one host computer, and when combining data from multiple hosts.
Unfortunately, memory addresses are only valid for one process.
To account for multiple processes on the same host, we combine the pointer value with the process ID (PID).
Then, to account for multiple hosts, we combine the pointer value and PID with a unique host ID obtained from the trace data.
This 3-tuple is thus unique across different processes and hosts.

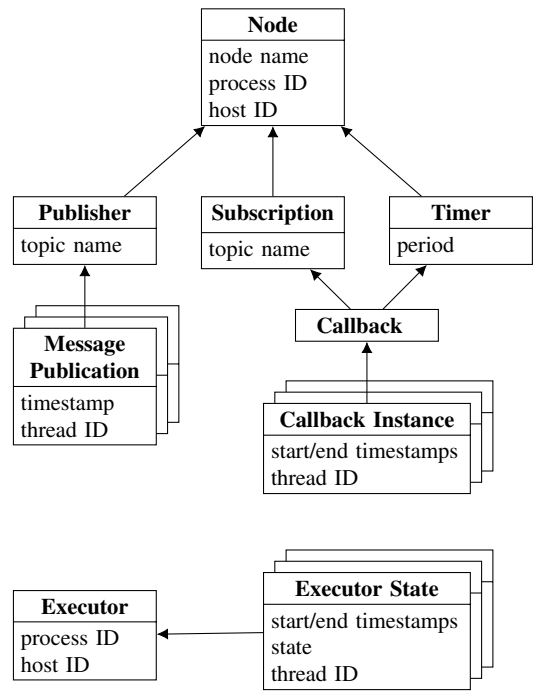
\begin{figure}[htb!]
\begin{center}

\renewcommand{\umlfillcolor}{white}
\renewcommand{\umldrawcolor}{black}

\tikzset{>={Latex[width=1.5mm,length=1.5mm]},
  baseTxt/.style = {scale=0.85},
  dupl/.style = {double copy shadow={shadow xshift=0.15cm, shadow yshift=0.15cm}}
}

\begin{tikzpicture}
  \begin{class}[baseTxt,text width=2cm]{Node}{0,0}
    \attribute{node name}
    \attribute{process ID}
    \attribute{host ID}
  \end{class}

  \begin{class}[baseTxt,text width=2cm]{Publisher}{-2.5,-2.5}
    \attribute{topic name}
  \end{class}
  \begin{class}[baseTxt,text width=2cm]{Subscription}{0,-2.5}
    \attribute{topic name}
  \end{class}
  \begin{class}[baseTxt,text width=2cm]{Timer}{2.5,-2.5}
    \attribute{period}
  \end{class}

  \begin{class}[baseTxt,dupl,text width=2cm]{Message Publication}{-2.5,-4.25}
    \attribute{timestamp}
    \attribute{thread ID}
  \end{class}
  \begin{class}[baseTxt,text width=2cm]{Callback}{1.25,-4}
  \end{class}
  \begin{class}[baseTxt,dupl,text width=3cm]{Callback Instance}{1.25,-5.25}
    \attribute{start/end timestamps}
    \attribute{thread ID}
  \end{class}

  \begin{class}[baseTxt,text width=2cm]{Executor}{-2.5,-7.74}
    \attribute{process ID}
    \attribute{host ID}
  \end{class}

  \begin{class}[baseTxt,dupl,text width=3cm]{Executor State}{1.25,-7.5}
    \attribute{start/end timestamps}
    \attribute{state}
    \attribute{thread ID}
  \end{class}

  \draw[->] (Publisher)           -- node[] {} (Node);
  \draw[->] (Subscription)        -- node[] {} (Node);
  \draw[->] (Timer)               -- node[] {} (Node);
  \draw[->] (Callback)            -- node[] {} (Subscription);
  \draw[->] (Callback)            -- node[] {} (Timer);
  \draw[->] (Message Publication) -- node[] {} (Publisher);
  \draw[->] (Callback Instance)   -- node[] {} (Callback);
  \draw[->] (Executor State)   -- node[] {} (Executor);
\end{tikzpicture}
 \end{center}
\vspacefiguretikz
\caption{
Information in the intermediate representation database.
Arrows represent relationships between objects and instances in the database.
For example, each message publication instance is linked to its corresponding publisher object, which is itself linked to its corresponding node.
}
\label{fig:data-model-model}
\end{figure}

In summary, as depicted in \cref{fig:data-model-model}, the resulting intermediate representation database contains information about all \ROStwo objects: nodes, publishers, subscriptions, timers, and executors.
It also contains all relevant instances: message publication instances, subscription and timer callback instances, and executor states over time.
This database can then be queried for analysis purposes.
As for services and actions, while they are not included in the model, the corresponding objects and request-response instances could easily be added, since they are similar to the current model.

\subsection{Implementation Details}

To build a database of raw trace data as an intermediate representation, as described in the previous section, we used Eclipse Trace Compass~\cite{tracecompass}, an open-source trace analysis framework.
Since traces collected from multiple computers usually do not have the same clock reference, the traces need to be synchronized for the combined data to be valid, time-wise.
Trace Compass can synchronize traces from distributed systems using network packet data collected from the kernel~\cite{jabbarifar2014liana}.
System clocks can also be synchronized directly using NTP~\cite{mills1991internet}.
The time synchronization method and its precision should of course be taken into consideration when extracting time-related information from the database.
We then use Trace Compass to perform our message flow analysis using information from the intermediate execution representation database: this is greatly simplified by using higher-level, preprocessed information (as shown in \cref{fig:data-model-model}) instead of raw trace data.

\section{Message Flow Analysis}\label{sec:analysis}

Our proposed message flow analysis builds a graph of the path of messages across a \ROStwo system using the information from the intermediate execution database, described in the previous section.
However, to achieve this, we must add more information to the database and combine multiple elements.
We must first track messages as they are sent over the network transport, to link a message being published by a publisher to the same message being received by one or more subscriptions (\cref{sec:analysis-transport-links}).
Then we add causal links, between messages that act as an input to a node, to the messages that are published by that node as the output (\cref{sec:analysis-causal-links}).
Finally, we put everything together to build the flow graph (\cref{sec:analysis-building-graph}).

\subsection{Transport Links}\label{sec:analysis-transport-links}

\begin{figure*}[htbp]
\centerline{\includegraphics[width=\textwidth,trim={0px 945px 2310px 0px},clip]{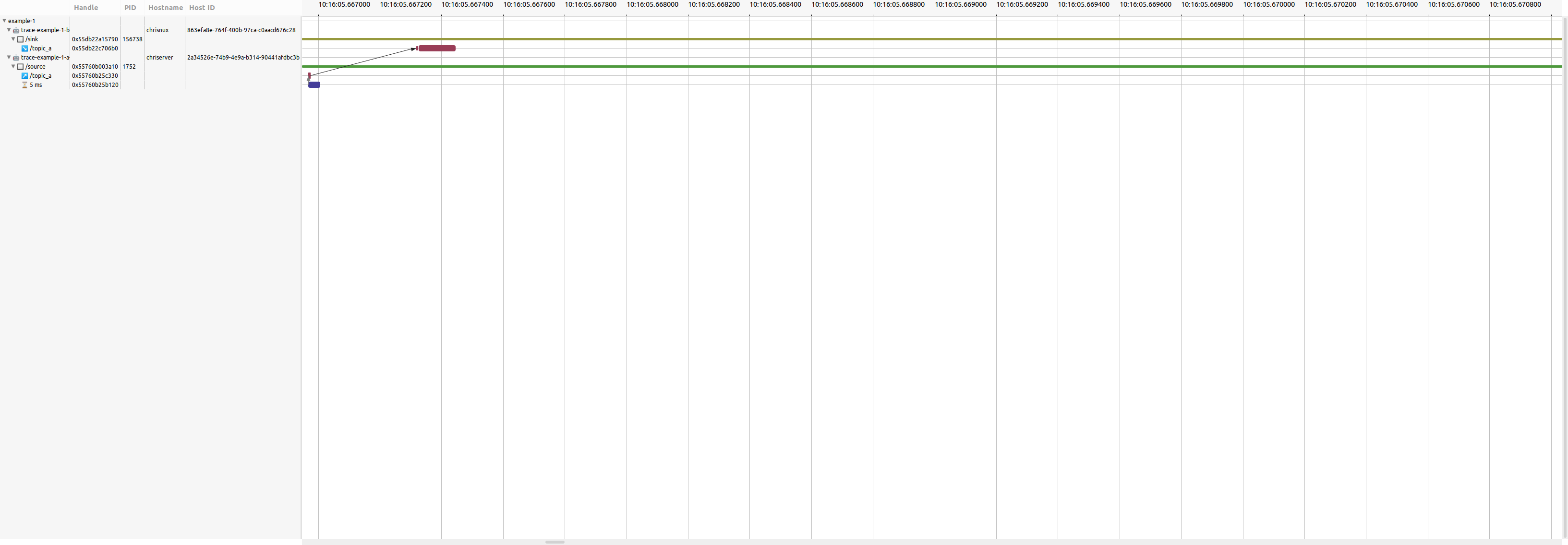}}
\vspacefigureimage
\caption{
Transport link example.
The tree structure on the left represents traces, with publishers, subscriptions, and timers under the nodes of each trace.
To the right of this, internal handles, PIDs, and host information are shown: this is the 3-tuple needed to uniquely identify \ROStwo objects (see \cref{sec:data-model-processing}).
Then, on the right is a time-based chart, which provides an abstract representation of the execution using time segments and arrows.
In this example, a 5~ms timer triggers a callback which publishes a message on \texttt{/topic\_a} under node \texttt{/source}.
This message is received by the \texttt{/topic\_a} subscription of node \texttt{/sink} on the other computer.
Next to timers and subscriptions, segments represent the duration of a specific callback instance, from beginning to end.
The smaller segment before the subscription callback segment represents the message being fetched (or \textit{taken}) from the underlying middleware, before it is provided to the callback instance.
For publishers, the segments represent the duration between the initial user-level publication call and the underlying DDS call.
The longer arrow between the \texttt{/topic\_a} publisher and subscription represents the transport link, i.e., the message going from the publisher to the subscription over the network.
The shorter arrow between the 5~ms timer callback and the \texttt{/topic\_a} publisher shows that the message was published during the timer callback.
}
\label{fig:analysis-transport-links-distributed}
\end{figure*}

Transport links associate a publication instance to the corresponding subscription callback instance on the other end.
These links are always one-to-many, since messages always originate from a single publisher but can be received by any number of subscriptions.

\ROStwo internally provides metadata for all received messages, including the GID of the source publisher and the timestamp of the time right before DDS sent the message over the network.
This information is collected on the subscription side using an instrumentation point in \texttt{rmw}.
The publisher GID is collected during initialization; the source timestamp is collected on the publisher side using DDS instrumentation, since this information is not made available to \ROStwo.
To uniquely identify messages and thus avoid collisions in case multiple publishers emit messages at the same time, we should use a combination of the publisher GID and the source timestamp.
Unfortunately, as of writing this, a bug in the implementation of the \texttt{rmw} interface for Cyclone DDS\footnotemark{} prevents us from relying on the GID.
\footnotetext{\href{https://github.com/ros2/rmw_cyclonedds/issues/377}{github.com/ros2/rmw\_cyclonedds/issues/377}}%
We therefore instead reduce the probability of collisions by combining the source timestamp with the topic name, which is known on both sides of the transport link.
Furthermore, unless the source clock has a higher granularity, collisions are unlikely, given that source timestamps have nanosecond-level precision.
These elements are all available from the intermediate execution representation database.

Collecting this low-level execution information, to track messages, allows our method to work transparently, i.e., without needing to modify a system to add fields to messages or rely on high-level tracking logic, unlike what is done by~\cite{nishimura2021raplet,witte2021inferred}.
More importantly, we expect the overhead to be much smaller, given the low overhead of the \rostwotracing instrumentation, as demonstrated in~\cite{bedard2022ros2tracing}.
\cref{fig:analysis-transport-links-distributed} shows an example of a transport link, with a subscription on one computer receiving a message from a publisher on another computer.
Given our technique for tracking messages and uniquely identifying \ROStwo objects (see \cref{sec:data-model-processing}), there is no difference between a transport link between two computers, and a transport link constrained to a single computer.

\subsection{Causal Message Links}\label{sec:analysis-causal-links}

For causal links, we define the causality of messages based on both time and value.
In direct cases, an output message is generated and published when a new input message is received and processed, thus linking the two messages.
In indirect cases, an input message is linked to an output message if the content of the former is used to generate the content of the latter, without any strict requirements on time.
Indeed, the link is not strictly time-related, since causal links can be asynchronous, as we will explain in the following.

\subsubsection{Direct Case}\label{sec:analysis-causal-link-direct}

\begin{figure}[htbp]
\centerline{\includegraphics[width=\columnwidth,trim={0px 175px 2550px 0px},clip]{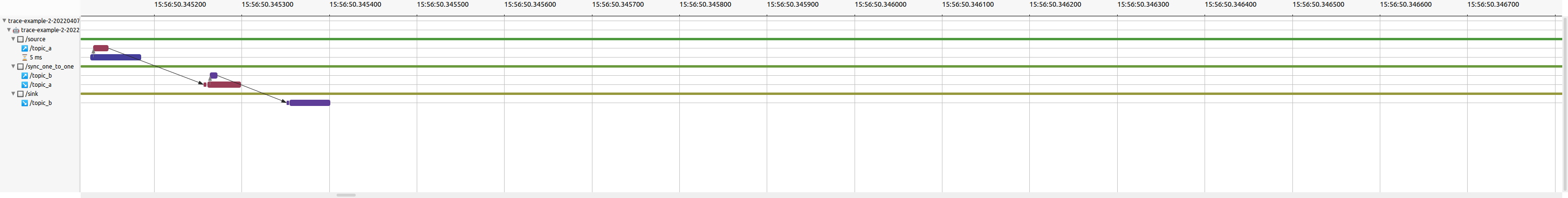}}
\vspacefigureimage
\caption{
Direct causal message link example.
A message is published on \texttt{/topic\_a} during a 5~ms timer callback by node \texttt{/source}.
The message is then received by the corresponding subscription under node \texttt{/sync\_one\_to\_one}.
During the subscription callback for that message, a message is published on \texttt{/topic\_b}, which is finally received by the corresponding subscription under node \texttt{/sink}.
The link between the input message and the output message is therefore a direct one-to-one causal link.
}
\label{fig:analysis-causal-links-direct}
\end{figure}

For the direct case, new messages are published on any number of topics directly during the subscription callback for a received message.
The input message is thus linked to all messages that are published between the start and end of the corresponding subscription callback instance on the same thread.
For example, by collecting execution information during runtime using \rostwotracing, we obtain the following trace events:
\begin{enumerate}
\item At timestamp $X$: start of callback on thread $T$ for message $I$ received by subscription $S$
\item At timestamp $Y$: publication of message $O$ with publisher $P$ on thread $T$
\item At timestamp $Z$: end of callback on thread $T$ for message $I$ received by subscription $S$
\end{enumerate}
With $X<Y<Z$, message $I$ received by subscription $S$ is directly linked to message $O$ published by publisher $P$, since it was published during the callback for message $I$ on the same thread $T$.
This direct causal link can therefore be inferred using the execution information collected with \rostwotracing; no user-level annotation is necessary for this case.
Since normal \ROStwo subscription callbacks only process a single message, the causal link for the direct case is strictly one-to-many.
Another example is shown in \cref{fig:analysis-causal-links-direct}, with a pipeline of three nodes and direct one-to-one causal links.
In this case, the message flow graph generated from this exchange would be visually identical, since there are no additional links to be found.

\subsubsection{Indirect Case}\label{sec:analysis-causal-link-indirect}

\begin{figure*}[t!]
\centering
\subfloat[
Periodic asynchronous causal message link.
All messages received by node \texttt{/periodic\_async\_n\_to\_m} are cached.
The periodic callback triggered by the 8~ms timer then uses those cached messages to compute and publish an output message on \texttt{/topic\_c}.
Therefore, the subscription callback of the input messages (\texttt{/topic\_a} and \texttt{/topic\_b}) are linked to the output message publication (\texttt{/topic\_c}).
These two periodic asynchronous links are shown in red.
]{%
\includegraphics[width=\columnwidth,trim={0px 915px 2450px 0px},clip]{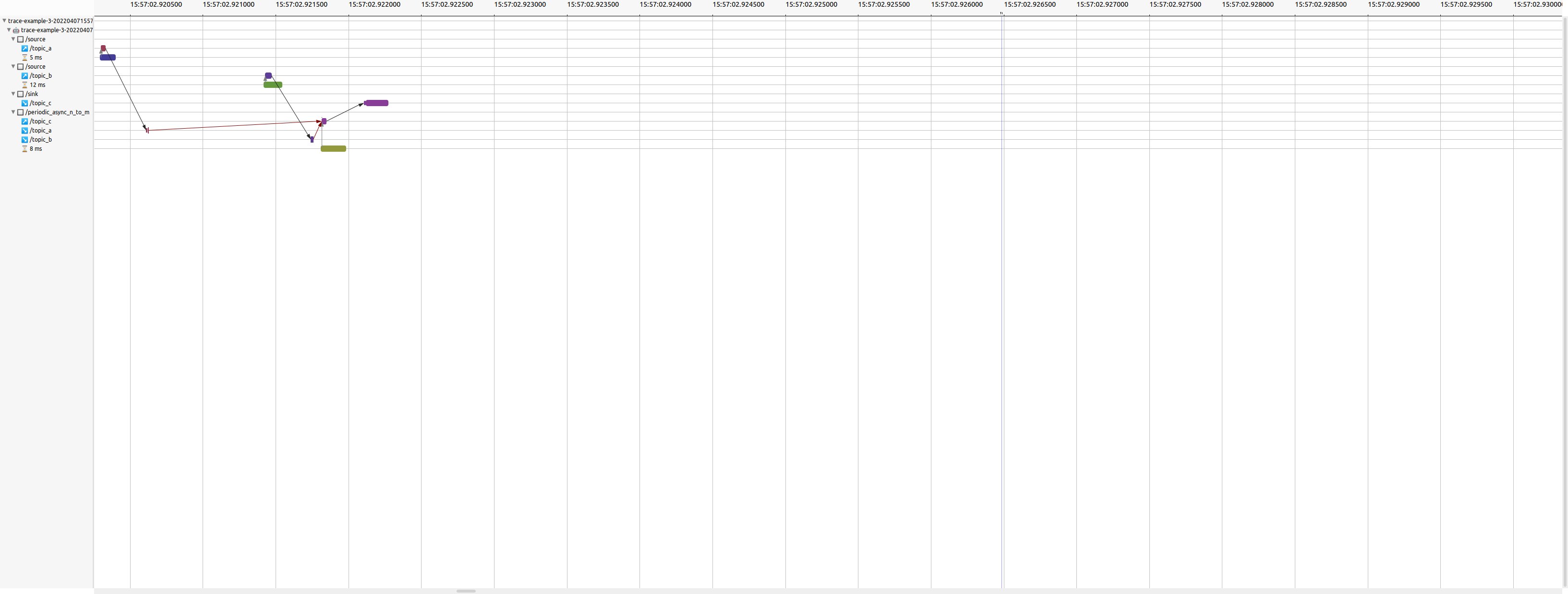}%
\label{fig:analysis-causal-links-periodic-async-flow}%
}
\hfil
\subfloat[
Partial synchronous causal message link.
Messages are received by node \texttt{/partial\_sync\_n\_to\_m}.
The first message (\texttt{/topic\_b}) is received and cached, since the other cache is empty.
However, when the second message (\texttt{/topic\_a}) is received, both messages are available, and are therefore used to compute and publish an output message on \texttt{/topic\_c}.
Therefore, the subscription callback of the first message (\texttt{/topic\_b}) is linked to the output message publication (\texttt{/topic\_c}) which happens during the subscription callback for the second message (\texttt{/topic\_a}).
]{%
\includegraphics[width=\columnwidth,trim={0px 850px 2450px 0px},clip]{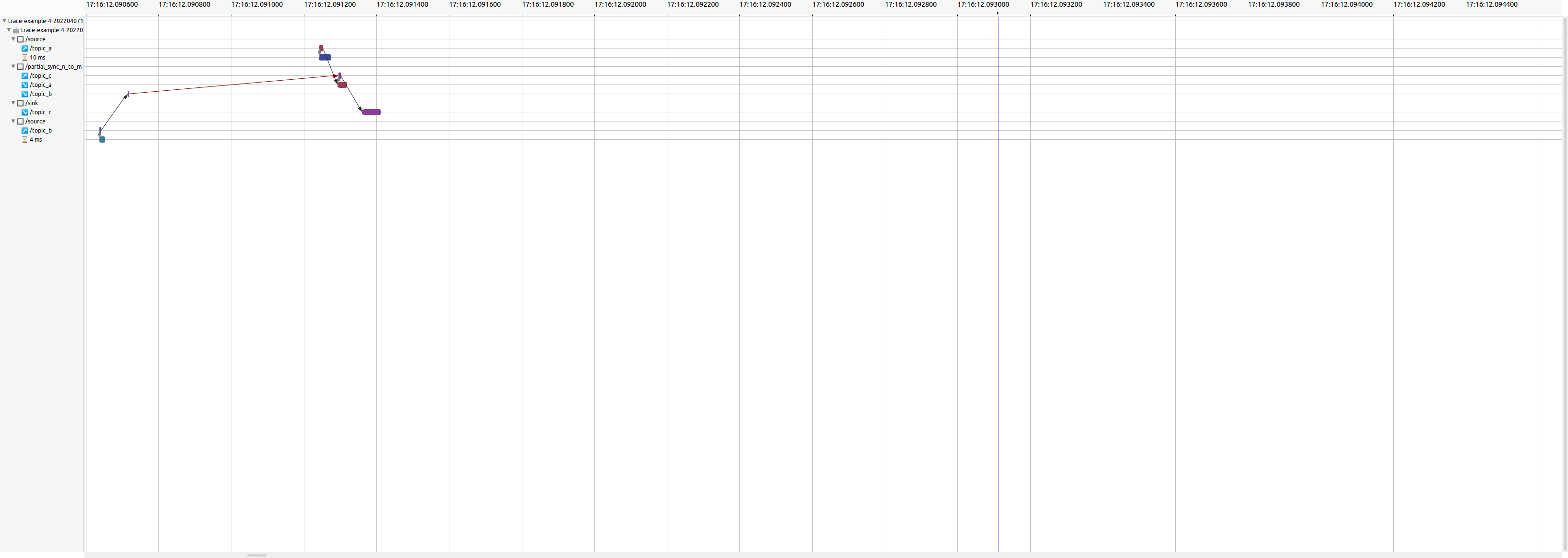}%
\label{fig:analysis-causal-links-partial-sync-flow}%
}
\caption{
Indirect causal message links examples: (a) periodic asynchronous and (b) partial synchronous.
In both examples, messages are published periodically using timers on \texttt{/topic\_a} and \texttt{/topic\_b} by two \texttt{/source} nodes.
These messages are then received by subscriptions under the \texttt{/periodic\_async\_n\_to\_m} and \texttt{/partial\_sync\_n\_to\_m} node, respectively.
An output message linked to the input messages is eventually published on \texttt{/topic\_c}.
}
\label{fig:analysis-causal-links-indirect}
\end{figure*}

For the indirect case, causal links are the result of user-level code, i.e., above the \ROStwo API.
We therefore cannot detect these causal links from the trace itself; users need to provide this application-specific information for the links to be detected.
We achieve this using simple annotations in the form of one-time tracepoints during the initialization phase, after the subscriptions and publishers have been created.
As shown in \cref{lst:user-annotation}, annotations simply contain the name of the message link type and the list of input subscriptions and output publishers as arrays along with their length.
By providing the connection between the input publishers and the corresponding output subscriptions and the message link type, indirect causal links between input messages and output messages can be inferred from the runtime execution data.
Annotation can thus be easily added to an existing system, without needing to modify the existing application logic or message structures as is done by~\cite{nishimura2021raplet,witte2021inferred}.

\begin{lstlisting}[
  caption={User annotation tracepoint example.},
  label={lst:user-annotation},
  language=C++,
  basicstyle=\ttfamily\small,
  frame=single
]
TRACEPOINT(
  message_link_type,
  (rclcpp::Subscription *) subscriptions,
  subscriptions_length,
  (rclcpp::Publisher *) publishers,
  publishers_length);
\end{lstlisting}
\vspacelisting

From studying real systems and the Autoware reference system~\cite{rtwgreferencesystem}, we define two types of causal links: periodic asynchronous link and partial synchronous link, both N-to-M, i.e., many-to-many.
With the periodic asynchronous link, messages are received from N topics and are cached.
A timer periodically triggers a callback during which the last message from each of the N caches is used to compute a result and then publish messages on M topics.
For the partial synchronous link, messages are received from N topics and are cached as well.
However, the result is conditionally computed during the subscription callback itself: if all N caches contain a message, a result is computed and published on M topics.
The caches are then reset so that at least one new message from each of the N input topics is received again before the next output.
Without the link annotation, the partial synchronous link would be similar to the direct case; however, it would only link one out of N real input messages.
Finally, while we only present two types of indirect causal links, more types could be identified.
If other link types are defined, i.e., with valid heuristics, then they can be easily added to our method.

With the information from the annotations and the timer and subscription callback, and message publications instances from the intermediate execution representation database, we can thus automatically infer indirect causal links between specific input and output messages.
Therefore, unlike~\cite{li2022autoware_perf}, users do not need to provide these links after the fact, since they are already in the trace data.
Furthermore, unlike the direct case, which is limited to a single input message per link, indirect causal links can be many-to-many, given their asynchronous nature.
\cref{fig:analysis-causal-links-periodic-async-flow} shows an example for the periodic asynchronous link, while \cref{fig:analysis-causal-links-partial-sync-flow} shows an example for the partial synchronous link.
In both cases, two input messages result in one output message.
However, the mechanics of the two causal links are different.
For the periodic asynchronous link, the output rate and delay between input and output depend on the period value for the timer.
For the partial synchronous link, the output rate only depends on the rate of the inputs.

\subsection{Building the Message Flow Graph}\label{sec:analysis-building-graph}

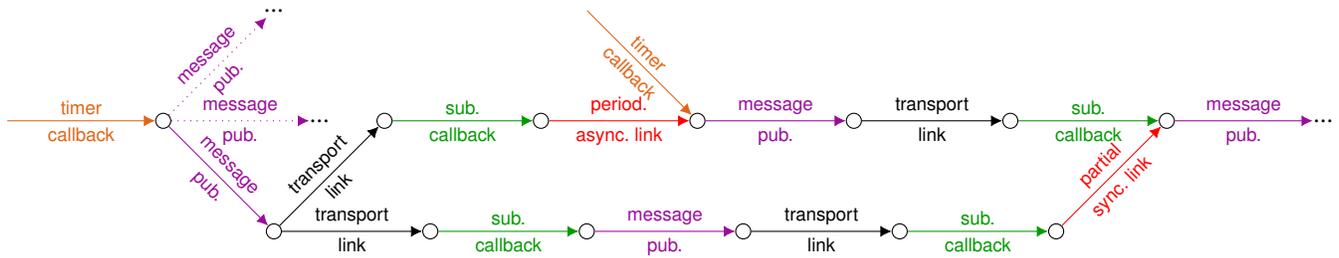
\begin{figure*}[htbp]
\begin{center}
\begin{tikzpicture}[align=center]
\tikzset{>={Latex[width=1.5mm,length=1.5mm]}}

\begin{scope}[
    every node/.style = {circle,draw,node distance=2.08cm,minimum size=0.20cm,inner sep=0pt}
]
    \node[minimum size=0cm] (a) {};
    \node[right of=a] (b) {};
    \node[below right of=b] (c) {};
    \node[right of=c] (d) {};
    \node[right of=d] (e) {};

    \node[above right of=b, draw=white] (f) {...};
    \node[right of=b, draw=white] (g) {...};

    \node[above right of=c] (h) {};
    \node[right of=h] (i) {};
    \node[right of=i] (j) {};
    \node[right of=j] (k) {};
    \node[right of=k] (l) {};
    \node[right of=l] (m) {};
    \node[right of=m, draw=white] (n) {...};

    \node[above left of=j, minimum size=0cm] (o) {};

    \node[right of=e] (p) {};
    \node[right of=p] (q) {};
    \node[right of=q] (r) {};
\end{scope}

\definecolor{darkorange}{RGB}{225,100,20}
\definecolor{darkpurple}{RGB}{155,10,155}
\definecolor{darkgreen}{RGB}{0,155,0}
\newcommand{\timercallback}  [3][]{\draw[->, #1, darkorange] (#2) -- node [baseTxt, above] {timer} node [baseTxt, below] {callback} (#3);}
\newcommand{\messagepub}     [3][]{\draw[->, #1, darkpurple] (#2) -- node [baseTxt, above] {message} node [baseTxt, below] {pub.} (#3);}
\newcommand{\transportlink}  [3][]{\draw[->, #1] (#2) -- node [baseTxt, above] {transport} node [baseTxt, below] {link} (#3);}
\newcommand{\subcallback}    [3][]{\draw[->, #1, darkgreen] (#2) -- node [baseTxt, above] {sub.} node [baseTxt, below] {callback} (#3);}
\newcommand{\periodasynclink}[3][]{\draw[->, #1, red] (#2) -- node [baseTxt, above] {period.} node [baseTxt, below] {async. link} (#3);}
\newcommand{\partialsynclink}[3][]{\draw[->, #1, red] (#2) -- node [baseTxt, above] {partial} node [baseTxt, below] {sync. link} (#3);}

\begin{scope}[
    baseTxt/.style = {scale=0.70, text centered, font=\sffamily, sloped}
]
    \timercallback{a}{b}
    \messagepub{b}{c}
    \transportlink{c}{d}
    \subcallback{d}{e}

    \messagepub[dotted]{b}{f}
    \messagepub[dotted]{b}{g}

    \transportlink{c}{h}
    \subcallback{h}{i}
    \periodasynclink{i}{j}
    \messagepub{j}{k}
    \transportlink{k}{l}
    \subcallback{l}{m}
    \messagepub{m}{n}

    \timercallback{o}{j}

    \messagepub{e}{p}
    \transportlink{p}{q}
    \subcallback{q}{r}
    \partialsynclink{r}{m}
\end{scope}

\end{tikzpicture}
 \end{center}
\vspacefiguretikz
\caption{
Simplified representation of a typical message flow graph, showing all edge types, each type with a specific color.
Edges are segments of the message flow, and their duration is their weight.
Vertices link one or more input edges to one or more output edges.
The third message (bottom) has two outgoing transport links, i.e., it is received by two subscriptions.
The first message (top) is processed by a subscription callback and put into a message cache.
This message has a periodic asynchronous causal link to an output message generated and published by a timer callback (above).
This last message is then received and processed by a subscription callback, which uses it along with another message linked by a partial synchronous causal link (below) to generate and publish a final message.
}
\label{fig:analysis-message-flow-graph}
\end{figure*}

Information from the intermediate execution representation database, including transport links and direct causal message links, can be displayed directly to provide a visual representation, as shown in \cref{fig:analysis-transport-links-distributed} and \cref{fig:analysis-causal-links-direct}.
We then need to use this information to build the message flow graph for a particular message, as selected by a user in the Trace Compass GUI.

As presented by Casini \etal~\cite{casini2019response} and used by~\cite{tang2020response,blass2021automatic,blass2021ros}, ROS computation graphs can be modeled as directed acyclic graphs (DAGs).
The flow of a message can therefore also be modeled as a DAG.
\cref{fig:analysis-message-flow-graph} shows a simplified version of a typical message flow graph.
Message flow graphs have a limited set of edge types.
Each edge type can be preceded and followed by a specific set of other edge types.
For example, transport link edges are preceded by message publication edges and followed by subscription callback edges.
Using this logic, and the different data sources presented in previous sections, the message flow graph can be constructed recursively, one edge at a time, going both forward and backward from the initial edge selected by the user.
For example, an output message will be linked to all input messages that are detected, possibly combining both direct and indirect causal links.
It is important to note that message flow graphs may be incomplete or invalid if links are not detected.
This could happen if the user does not correctly annotate indirect causal links, or if the traced system contains types of indirect causal links that are not supported.

Unlike~\cite{bedard2019messageflow}, which assumed for simplification purposes that the message flow graph is a directed graph that only contains one-to-one links, our method supports one-to-many transport links and many-to-many causal message links.
Furthermore, unlike~\cite{bedard2019messageflow} again, our method also builds the message flow graph both forward and backward from the initial element.
This can be useful when analyzing traces from a \ROStwo system: building a message flow graph starting from one of the roots of the \ROStwo computation DAG will be different from a graph built starting from the leaf of a computation DAG, even if the resulting message flow graphs intersect.
The initial segment from which to build the message flow graph can thus be chosen depending on the user needs.

Finally, there is one exception when modeling ROS computational graphs as DAGs.
As explained by Blass \etal~\cite{blass2021automatic}, the \texttt{/tf} topic is a special topic.
It is used to communicate information about relationships between coordinate frames (\textit{transforms}).
All nodes that need and provide this information both subscribe and publish to the same topic, thus seemingly creating a loop in the graph model.
However, \texttt{/tf} messages have a field which identifies the two coordinate frames to which the transform message applies.
Therefore, a node publishing a \texttt{/tf} message might also receive that same message; however, it will not be used.
As Blass \etal~\cite{blass2021automatic} do, we can detect the transport links for \texttt{/tf} with the same node as the source and destination, and remove them, since we assume that these messages are not meant for the originating node itself, and will not be used by it.
However, this does not actually cause loops in our message flow graph implementation.
Indeed, the subscriptions to the \texttt{/tf} topic are special subscriptions that are managed by \ROStwo: messages are received and put into a buffer, the content of which is used eventually by the user.
This application-level link is not detected; therefore, our method does not detect any message flow segments after subscription callbacks for \texttt{/tf} messages.

\section{Executor State Visualization}\label{sec:executor}

\begin{figure}[htbp]
\centerline{\includegraphics[width=\columnwidth,trim={0px 948px 2575px 0px},clip]{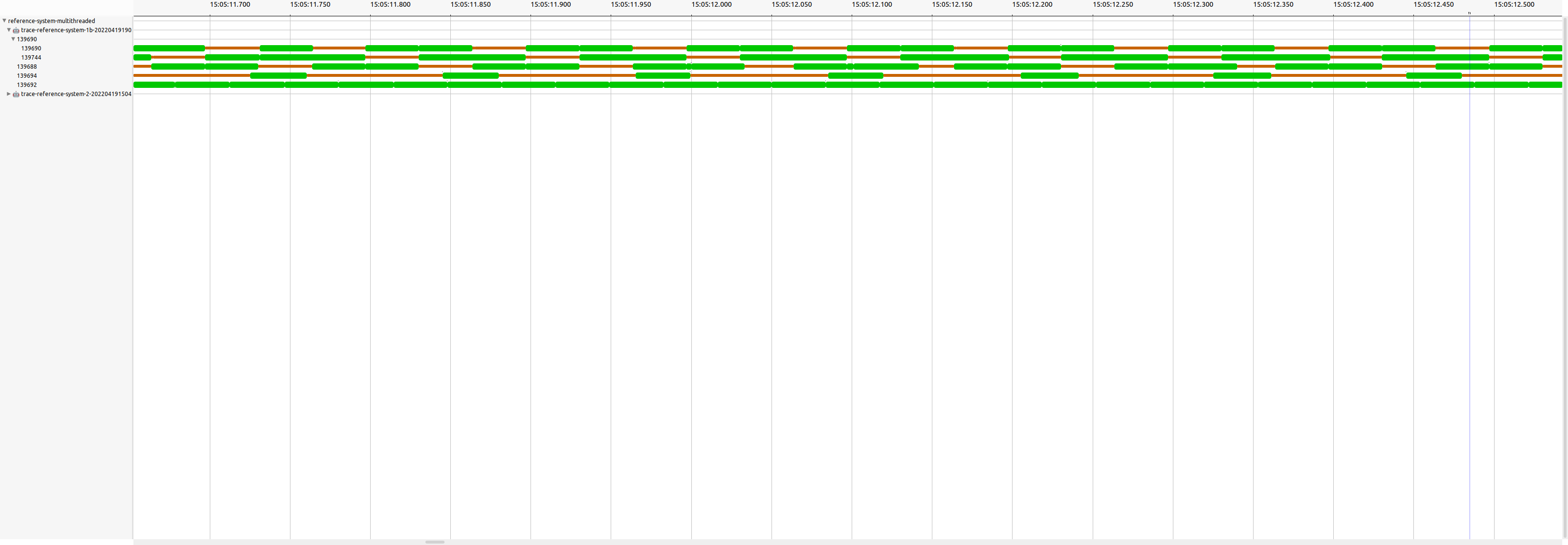}}
\vspacefigureimage
\caption{
Visualization of the state of executor instances for each process over time.
The tree structure on the left lists IDs of processes under each trace.
For multi-threaded executor instances, individual thread IDs are listed under the process ID.
Green segments represent execution instances (e.g., timer callback, subscription callback).
Orange segments indicate that the executor is waiting for new events.
Red segments represent internal executor processing, although they are too small to be visible for this time range.
}
\label{fig:executor-visualization}
\end{figure}

Using information from the intermediate representation database, we also build a visualization of the state of the executor over time.
As explained in \cref{sec:related-work-executor}, the \ROStwo executor is responsible for fetching new messages from the middleware and executing the user-provided subscription and timer callbacks.
We split the executor runtime into three distinct states that repeat over time.
In the first state, the executor waits for new events, such as new messages from the underlying middleware or timers that are ready to execute.
Then, in the second state, the executor does internal processing to select the subscription or timer that will be executed, which represents pure executor overhead.
This is an open problem in \ROStwo, along with suboptimal scheduling policies.
Finally, in the third state, the executor executes the corresponding user-provided subscription or timer callback.
\cref{fig:executor-visualization} shows an example of the resulting executor visualization over time.
For a given executor instance, orange means that the executor has nothing to process, while green means that it is busy executing callbacks.
This goes alongside the message flow analysis presented in \cref{sec:analysis}, since it can help explain message processing delays: even if a message is received, the corresponding subscription callback might wait for some time until the executor instance executes it.
This visualization can therefore be used to compare different executor designs and configurations to study and address the aforementioned open problems.

\section{Experiments}\label{sec:experiments}

We first apply our proposed method to a synthetic system and then apply it to a real system.
These experiments demonstrate how our technique can be used to analyze \ROStwo itself, as well as application-level logic, for performance optimization purposes.
The code and instructions for these experiments are available in our repository.

\subsection{Autoware Reference System}\label{sec:experiments-reference-system}

\begin{figure*}[htbp]
\centering
\subfloat[
End-to-end message flow graph.
The initial root of the message flow graph is a 100~ms timer callback instance which publishes a \texttt{/FrontLidarDriver} message, while the main leaves of the graph are \texttt{/VehicleInterface} messages received by the \texttt{VehicleDBWSystem} node.
The message flow graph includes direct and indirect causal links, including both periodic asynchronous and partial synchronous links.
]{%
\makebox[\textwidth]{%
\includegraphics[width=\ifx\outputstyle\elsevier0.85\else0.95\fi\textwidth,trim={0px 40px 2175px 0px},clip]{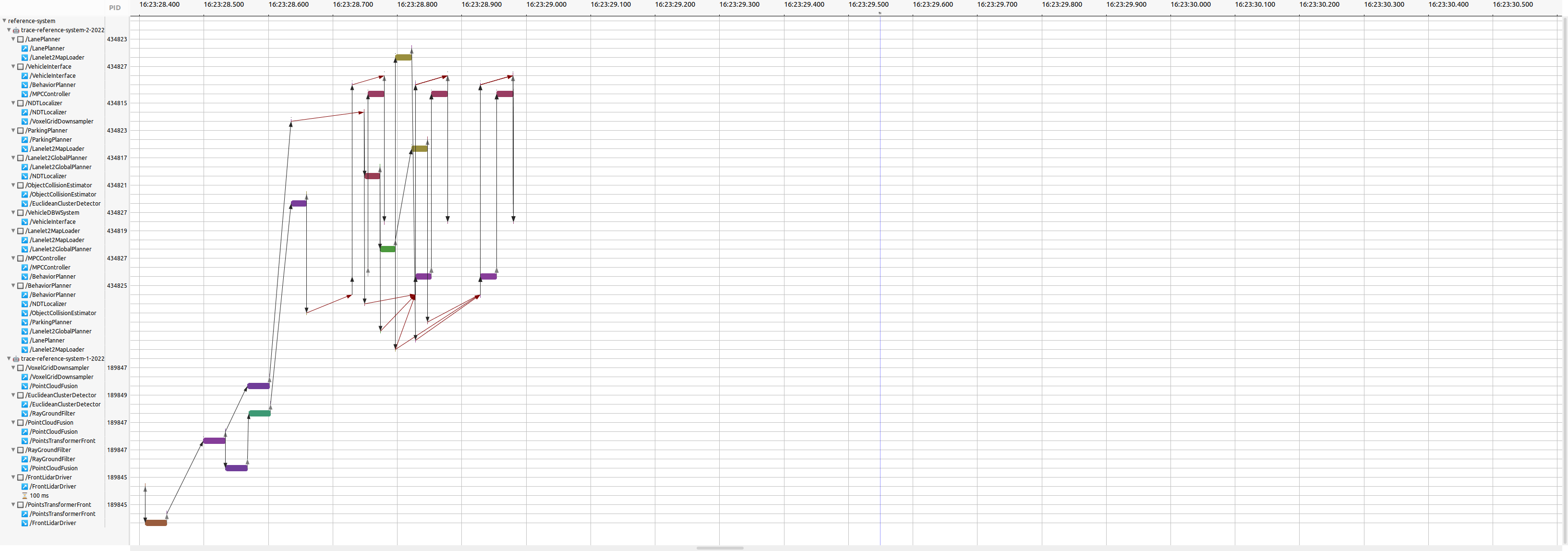}%
\label{fig:experiment-1-message-flow}%
}}
\hfil
\subfloat[
State of all executor instances over time for the same time range as (a).
Executor instances of the first host (lower half) are busier than the executor instances of the second host (upper half).
]{%
\makebox[\textwidth]{%
\includegraphics[width=\ifx\outputstyle\elsevier0.85\else0.95\fi\textwidth,trim={0px 90px 2175px 0px},clip]{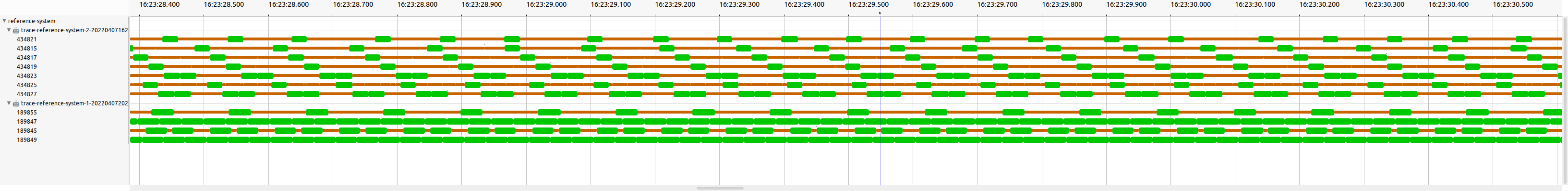}%
\label{fig:experiment-1-executor-state}%
}}
\caption{
Autoware reference system (a) message flow result example and (b) executor state for the same time range.
}
\label{fig:experiment-1-flow-and-executor}
\end{figure*}

\begin{figure*}[htbp]
\centering
\subfloat[
Partial message flow graph; segments for the second host are hidden.
Callbacks for the same \texttt{/PointCloudFusion} message received by the \texttt{/RayGroundFilter} and \texttt{/VoxelGridDownsampler} nodes are executed concurrently.
]{%
\includegraphics[width=\textwidth,trim={0px 140px 2400px 0px},clip]{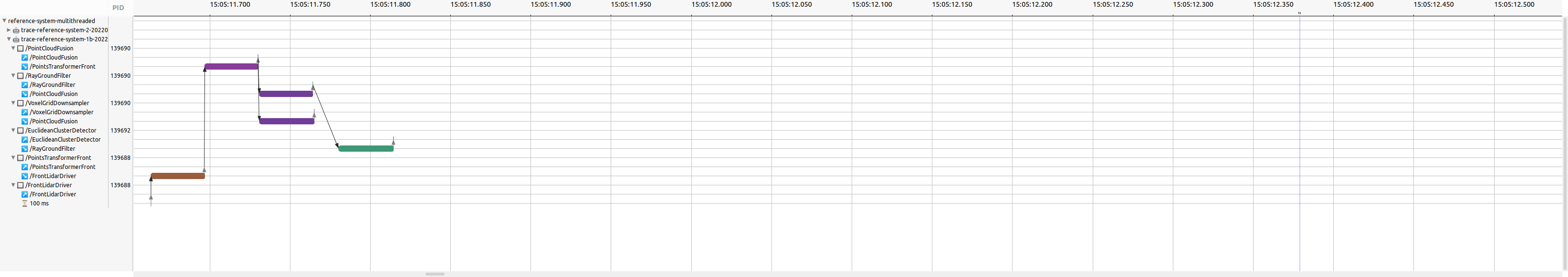}%
\label{fig:experiment-1-multithreaded-message-flow}%
}
\hfil
\subfloat[
State of executor instances for the first host over time for the same time range as (a).
Process 139690 has two executor threads (139690 and 139744), while other processes use single-threaded executors.
]{%
\includegraphics[width=\textwidth,trim={0px 925px 2400px 0px},clip]{figures/6_experiment_1_multithreaded_executor.png}%
\label{fig:experiment-1-multithreaded-executor-state}%
}
\caption{
Autoware reference system (a) message flow and (b) executor state for the same time range, showing the impact of a multi-threaded executor instance.
}
\label{fig:experiment-1-multithreaded-flow-and-executor}
\end{figure*}

For our first experiment, we use a reference system~\cite{rtwgreferencesystem} with a synthetic computation graph based on Autoware~\cite{kato2018autoware,kato2015open}, a ROS-based autonomous driving stack.
The nodes and topics are based the Autoware system; however, all messages have the same type, and computation is replaced with a processing-intensive task, in order to consume CPU time.
The computation graph has multiple inputs and outputs, i.e., sensor data and vehicle commands or secondary visualization outputs.
It uses all types of causal message links, as defined in \cref{sec:analysis-causal-links}.
However, it specifically uses the periodic asynchronous link in a many-to-one configuration, and the partial synchronous link in a two-to-one configuration.

We split the nodes defined in the reference system into multiple executables and split those executables into two launch files.
Each launch file is run on a specific host, with the two hosts being on the same network.
We set the launch files to configure the LTTng tracer using \rostwotracing and enable all \ROStwo and DDS tracepoints.
Furthermore, we enable network-related events, in order to synchronize the traces using~\cite{jabbarifar2014liana}.

\cref{fig:experiment-1-flow-and-executor} shows the entire message flow graph starting from one of the lidar drivers, which is one of the roots of the computation graph, along with the state of the executors over time.
As displayed in \cref{fig:experiment-1-message-flow}, the initial \texttt{/FrontLidarDriver} message results in three separate \texttt{/VehicleInterface} messages to the \texttt{/VehicleDBWSystem} node.
This is due to the caching and asynchronous nature of the indirect causal links, which results in one-to-many or many-to-many links, as explained in \cref{sec:analysis-causal-link-indirect}.
The end-to-end latency ranges from 370~ms for the first message to 571~ms for the third message.

Furthermore, as displayed in \cref{fig:experiment-1-executor-state}, a secondary visualization shows the state of all executor instances over time.
In this experiment, we only use the default single-threaded executor, which means that timer and subscription callbacks within a single process can only be processed one at a time.
We can see that some executor instances are busier than others; if executors are too busy, there can be a greater delay between message reception and processing.
Depending on subscription options, old messages could be dropped, which wastes the CPU time used for publishing those dropped messages, thus resulting in a generally poor performance optimization.
For example, unlike all executor instances under \texttt{trace-reference-system-2}, the executor instance for process 189847 is always busy, which could explain why the callback for the \texttt{/PointsTransformerFront} message under the \texttt{/PointCloudFusion} node happens long after the message was published.
Also, the \texttt{/VoxelGridDownsampler}, \texttt{/PointCloudFusion}, and \texttt{/RayGroundFilter} nodes are all on the same process and thus share the same single-threaded executor instance.
The callbacks for the same \texttt{/PointCloudFusion} message under two different nodes thus cannot be processed at the same time.
This directly affects the end-to-end latency, as our method helps highlight.

In this case, the computation graph distribution could be improved: nodes could be split over more processes, and better executor designs could be used.
For instance, as demonstrated in \cref{fig:experiment-1-multithreaded-flow-and-executor}, to allow the callbacks of the \texttt{/VoxelGridDownsampler} and \texttt{/RayGroundFilter} nodes to run simultaneously, we can use a multi-threaded executor with 2 threads.
As shown in \cref{fig:experiment-1-multithreaded-executor-state}, there are two executor threads for the corresponding process (139690 and 139744).
The subscription callbacks can then run concurrently, as shown in \cref{fig:experiment-1-multithreaded-message-flow}.
The end-to-end latency is hence reduced from 370~ms in the previous example to 298~ms in this example.
Our method can therefore be used to compare or study the impact of proposed executor designs in order to address the open problems summarized in \cref{sec:related-work}.

\subsection{RTAB-Map}\label{sec:experiments-rtabmap}

\begin{figure*}[htbp]
\centering
\subfloat[
Display of intermediate execution representation data: subscription callbacks, message publications, transport links (black arrows), and direct causal links (gray arrows).
A camera driver node (\texttt{/camera/camera}) and odometry node (\texttt{/rgbd\_odometry}) run on the the first host (top half) along with a transform listener node (see \cref{sec:analysis-building-graph}).
The RTAB-Map (\texttt{/rtabmap}) and rviz (\texttt{/rviz}) nodes run on the second host (bottom half) along with two transform listener nodes.
]{%
\includegraphics[width=\textwidth,trim={0px 60px 2050px 0px},clip]{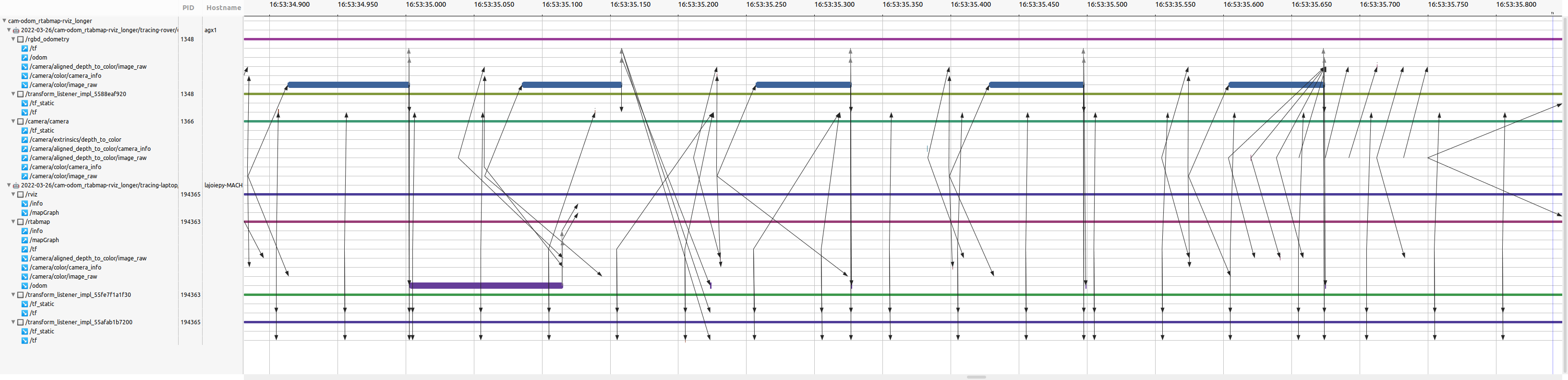}%
\label{fig:experiment-2-messages}%
}
\hfil
\subfloat[
End-to-end message flow graph for the main RTAB-Map computation pipeline, which goes through, in order: camera driver node (\texttt{/camera/camera}), odometry node (\texttt{/rgbd\_odometry}), RTAB-Map node (\texttt{/rtabmap}), and rviz node (\texttt{/rviz}).
The graph was generated starting from the subscription callback for a \texttt{/mapGraph} message received by the \texttt{/rviz} node, which is the last segment of the computation pipeline.
Therefore, the message flow only goes backward from there to the root \texttt{/camera/color/image\_raw} message published by the \texttt{/camera/camera} node on the other host.
Looking at \cref{fig:experiment-2-messages}, we know that generating the message flow graph starting from the initial \texttt{/camera/color/image\_raw} message would result in a message flow graph with one-to-many links, similar to \cref{fig:experiment-1-message-flow}.
]{%
\includegraphics[width=\textwidth,trim={0px 260px 2050px 0px},clip]{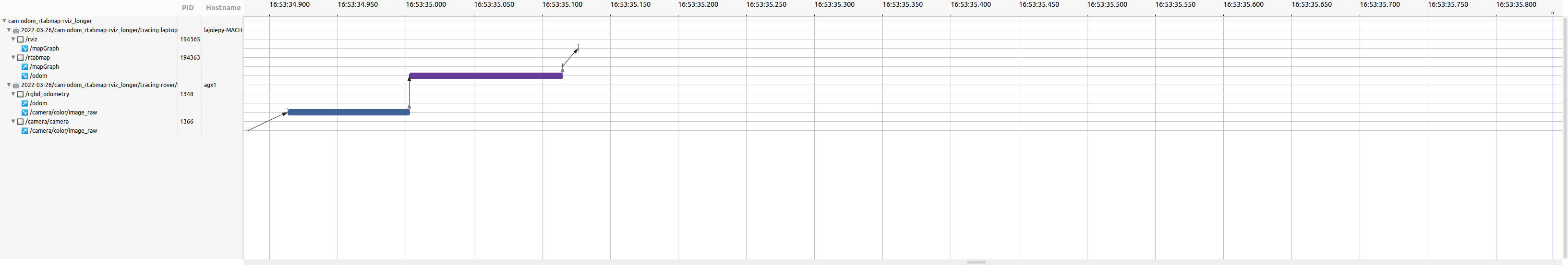}%
\label{fig:experiment-2-message-flow}%
}
\caption{
RTAB-Map (a) callback instances and message publications along with transport and direct links, and (b) message flow result for the main computation pipeline for the same time range.
}
\label{fig:experiment-2-messages-and-flow}
\end{figure*}

For our second experiment, we distribute the RTAB-Map~\cite{labbe2019rtab} simultaneous localization and mapping (SLAM) system over two computers and trace it.
One host, an AgileX Scout Mini equipped with an NVIDIA Jetson AGX Xavier and an Intel RealSense D435i camera, runs the camera driver node and odometry node.
The other host, a laptop on the same wireless network, runs the main SLAM node and rviz to visualize the resulting map.
To obtain synchronized traces, we synchronize the clocks of the two hosts using NTP~\cite{mills1991internet}.

A section of the trace is represented in \cref{fig:experiment-2-messages-and-flow}.
\cref{fig:experiment-2-messages} shows all callback instances, message publications, transport links, and direct links from the intermediate execution representation database for this time range.
\cref{fig:experiment-2-message-flow} shows an end-to-end message flow graph for the main computation pipeline.
Using Trace Compass, we find that the end-to-end latency is 242.3~ms; the duration of the first subscription callback (odometry computation) is 89.5~ms, and the duration of the second subscription callback (RTAB-Map) is 112.5~ms.
As seen shortly after the 16:53:35.000 time mark under the \texttt{/rgbd\_odometry} node, a \texttt{/tf} message is published during a subscription callback instance.
As mentioned in \cref{sec:analysis-building-graph}, the message is received by a special subscription, a transform listener, under the same process (PID 1348).
However, this \texttt{/tf} message is actually only intended for the two transform listeners on the other hosts (PIDs 194363 and 194365), and does not cause a loop since there are no further segments after these \texttt{/tf} messages are received.
Our method could be improved to model and detect indirect causal links after \texttt{/tf} messages.

\section{Runtime Overhead Evaluation}\label{sec:overhead-evaluation}

Since runtime overhead should be minimal to avoid perturbing an application~\cite{gregg2020systems}, we also evaluate the overhead of execution data collection.
In previous work, we demonstrated that \rostwotracing~\cite{bedard2022ros2tracing} introduces a mean end-to-end latency overhead of 0.0033~ms for a single message publication (i.e., publisher to subscription).
Since the instrumentation proposed in~\cite{bedard2022ros2tracing} includes 10 tracepoints in the publish-subscribe hot path, this is comparable to a runtime cost per LTTng userspace tracepoint of 158~ns, as measured by~\cite{gebai2018survey}.
Depending on the DDS implementation, our proposed method adds either 2 or 3 additional tracepoints to the hot path.
However, the systems presented in \cref{sec:experiments} include between 3 and 8 message transport instances, and have end-to-end latencies ranging from 240~ms to 370~ms.
Furthermore, as discussed in~\cite{bedard2022ros2tracing}, the combination of a high-level \ROStwo scheduler on top of the OS scheduler and networking stack introduces a lot of variability.
Therefore, we expect the end-to-end latency overhead for real applications to be small, especially when compared to the absolute latency.

\begin{figure}[htbp]
\centerline{\includegraphics[width=\columnwidth]{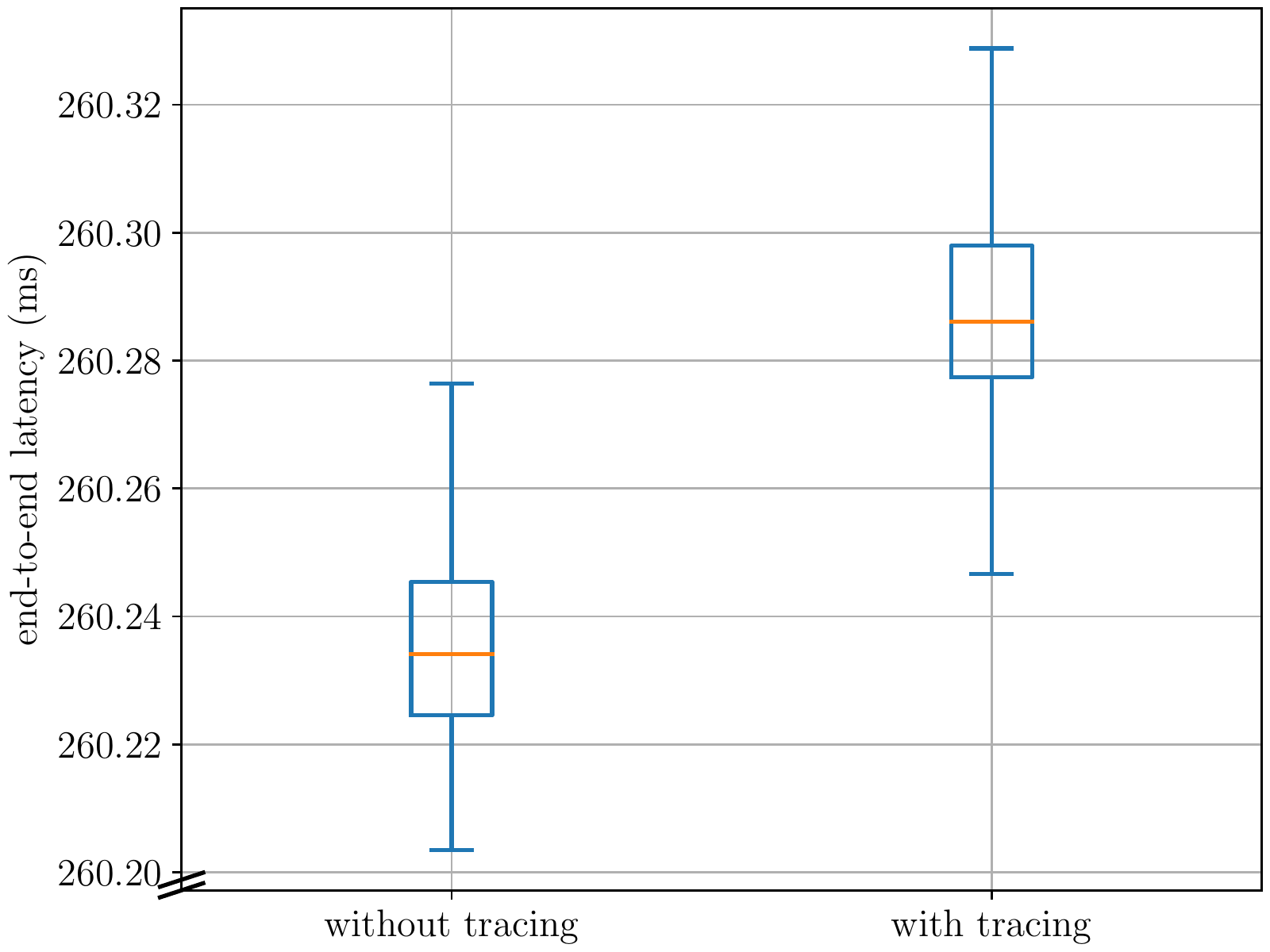}}
\vspacefigureimage
\caption{
End-to-end latency comparison, without tracing (left) and with tracing (right).
}
\label{fig:overhead-evaluation-results}
\end{figure}

We create a computation graph similar to the experiments in \cref{sec:experiments}, with 5 one-to-one message transport instances and an expected end-to-end latency of approximately 260~ms.
We use an Intel i7-3770 (3.40 GHz) 4-core CPU, 8 GB RAM system with Ubuntu 20.04.2, and disable power-saving features.
We run the computation graph at 10 Hz for 20 minutes first without and then with tracing to compare the runtime impact of tracing on the end-to-end latency.
By comparing the latencies for each case, shown in \cref{fig:overhead-evaluation-results}, we obtain a difference of means of 0.1597~ms and a difference of medians of 0.0521~ms.
This end-to-end latency overhead is small compared to a total latency of 260~ms; we therefore consider it suitable for real applications.
Furthermore, this value is within an order of magnitude of overhead values extrapolated from results by~\cite{bedard2022ros2tracing} and~\cite{gebai2018survey}, respectively 0.0215~ms and 0.0103~ms.
Finally, as mentioned previously, we expect this overhead to be less noticeable -- and challenging to actually measure -- on more complex \ROStwo systems.

\section{Future Work}\label{sec:future-work}

Many improvements and additions could be made to our proposed message flow analysis method and executor state visualization.

First, our method can easily be extended with other indirect causal links.
Moreover, transport links could be split into actual network transport time and a time delay between message reception by DDS and processing by the executor.
As mentioned in \cref{sec:experiments-reference-system}, it would help highlight delays in callback executions when the executor is busy processing other callbacks.
This would require additional instrumentation in the underlying DDS middleware.
Other executor types, such as the multi-threaded executor or other executors presented in previous work~\cite{lange2018callback,choi2021picas,staschulat2020rclc,staschulat2021budget}, could also be instrumented and supported for the executor state visualization.
Additionally, as mentioned in \cref{sec:analysis-building-graph} and shown in \cref{sec:experiments-rtabmap}, special subscriptions like the transform listener could be supported to be able to detect valid message flow segments resulting from received \texttt{/tf} messages.
Finally, it would also be interesting to extend this work to include \ROStwo services and actions.

Furthermore, our message flow analysis method could be further extended into a critical path analysis~\cite{yang1988critical}.
Fundamentally, indirect causal links (shown in red in \cref{fig:analysis-causal-links-indirect}, \cref{fig:analysis-message-flow-graph}, and \cref{fig:experiment-1-message-flow}) are wait intervals.
Indeed, as mentioned in \cref{sec:analysis-causal-link-indirect}, the duration of a periodic asynchronous link depends on the period of the timer, and the duration of a partial synchronous link depends on the other input messages.
These wait segments could thus be recursively replaced with the actual cause of the wait, as is done by Giraldeau and Dagenais~\cite{giraldeau2015wait} using kernel-level wait primitives for wait dependencies across a distributed system.

Similarly, the message flow graph could be augmented with other information.
For example, as mentioned in \cref{sec:data-model-processing}, the duration of a message publication call can be broken down into \texttt{rclcpp}, \texttt{rcl}, and DDS time.
The required information is already collected using \rostwotracing and available in the intermediate execution representation database.
Finally, other metrics, such as message publication or reception rate and executor usage over time, could be extracted from the database and displayed alongside our proposed visualizations.

\section{Conclusion}\label{sec:conclusion}

In conclusion, modern robotic systems are built as distributed computation graphs, using the publish-subscribe paradigm and frameworks such as \ROStwo.
However, there are open problems with the higher-level scheduling of tasks performed by the \ROStwo executor, which can affect performance.

We presented a low-overhead method for extracting and visualizing the flow of a message across a distributed \ROStwo system.
Our novel approach can detect exchanges of messages across distributed systems, and also introduces simple annotations for indirect causal message links.
This is achieved without needing to modify the applications themselves.
While we only define two types of indirect causal links, our work can be extended to consider other types of message links as well.
This would help ensure the validity of the message flow graphs generated for more complex systems.
Combined with a visualization of the state of the executor instances over time, our work is useful for optimizing both application layers and \ROStwo itself.

Finally, the underlying intermediate execution representation data can be leveraged for further analyses.
Furthermore, the message flow graph can also be extended with more information, and can be expanded into a critical path analysis, by recursively resolving wait dependencies resulting from indirect causal links.
 
\bibliographystyle{IEEEtran}

\begin{thebibliography}{10}
\providecommand{\url}[1]{#1}
\csname url@rmstyle\endcsname
\providecommand{\newblock}{\relax}
\providecommand{\bibinfo}[2]{#2}
\providecommand\BIBentrySTDinterwordspacing{\spaceskip=0pt\relax}
\providecommand\BIBentryALTinterwordstretchfactor{4}
\providecommand\BIBentryALTinterwordspacing{\spaceskip=\fontdimen2\font plus
\BIBentryALTinterwordstretchfactor\fontdimen3\font minus
  \fontdimen4\font\relax}
\providecommand\BIBforeignlanguage[2]{{%
\expandafter\ifx\csname l@#1\endcsname\relax
\typeout{** WARNING: IEEEtran.bst: No hyphenation pattern has been}%
\typeout{** loaded for the language `#1'. Using the pattern for}%
\typeout{** the default language instead.}%
\else
\language=\csname l@#1\endcsname
\fi
#2}}

\bibitem{lajoieTowards2022}
P.-Y. Lajoie, B.~Ramtoula, F.~Wu, and G.~Beltrame, ``Towards collaborative
  simultaneous localization and mapping: a survey of the current research
  landscape,'' \emph{Field Robotics}, vol.~2, no.~1, pp. 971--1000, 2022.

\bibitem{Dewangan2017}
R.~K. Dewangan, A.~Shukla, and W.~W. Godfrey, ``Survey on prioritized multi
  robot path planning,'' in \emph{2017 IEEE International Conference on Smart
  Technologies and Management for Computing, Communication, Controls, Energy
  and Materials (ICSTM)}, 2017, pp. 423--428.

\bibitem{yan2013}
\BIBentryALTinterwordspacing
Z.~Yan, N.~Jouandeau, and A.~A. Cherif, ``A survey and analysis of multi-robot
  coordination,'' \emph{International Journal of Advanced Robotic Systems},
  vol.~10, no.~12, p. 399, 2013. [Online]. Available:
  \url{https://doi.org/10.5772/57313}
\BIBentrySTDinterwordspacing

\bibitem{macenski2022ros2}
S.~Macenski, T.~Foote, B.~Gerkey, C.~Lalancette, and W.~Woodall, ``Robot
  operating system 2: Design, architecture, and uses in the wild,''
  \emph{Science Robotics}, vol.~7, no.~66, p. eabm6074, 2022.

\bibitem{quigley2009ros}
M.~Quigley, K.~Conley, B.~Gerkey, J.~Faust, T.~Foote, J.~Leibs, R.~Wheeler, and
  A.~Y. Ng, ``Ros: an open-source robot operating system,'' in \emph{ICRA
  workshop on open source software}, vol.~3, no. 3.2.\hskip 1em plus 0.5em
  minus 0.4em\relax Kobe, Japan, 2009, p.~5.

\bibitem{lange2018callback}
\BIBentryALTinterwordspacing
R.~Lange, ``Callback-group-level executor for ros 2,'' in \emph{ROSCon Madrid
  2018}.\hskip 1em plus 0.5em minus 0.4em\relax Open Robotics, September 2018.
  [Online]. Available: \url{https://vimeo.com/292707644}
\BIBentrySTDinterwordspacing

\bibitem{choi2021picas}
H.~Choi, Y.~Xiang, and H.~Kim, ``Picas: New design of priority-driven
  chain-aware scheduling for ros2,'' in \emph{2021 IEEE 27th Real-Time and
  Embedded Technology and Applications Symposium (RTAS)}.\hskip 1em plus 0.5em
  minus 0.4em\relax IEEE, 2021, pp. 251--263.

\bibitem{staschulat2020rclc}
J.~Staschulat, I.~L{\"u}tkebohle, and R.~Lange, ``The rclc executor:
  Domain-specific deterministic scheduling mechanisms for ros applications on
  microcontrollers: work-in-progress,'' in \emph{2020 International Conference
  on Embedded Software (EMSOFT)}.\hskip 1em plus 0.5em minus 0.4em\relax IEEE,
  2020, pp. 18--19.

\bibitem{staschulat2021budget}
J.~Staschulat, R.~Lange, and D.~N. Dasari, ``Budget-based real-time executor
  for micro-ros,'' \emph{arXiv preprint arXiv:2105.05590}, 2021.

\bibitem{nishimura2021raplet}
K.~Nishimura, T.~Ishikawa, H.~Sasaki, and S.~Kato, ``Raplet: Demystifying
  publish/subscribe latency for ros applications,'' in \emph{2021 IEEE 27th
  International Conference on Embedded and Real-Time Computing Systems and
  Applications (RTCSA)}.\hskip 1em plus 0.5em minus 0.4em\relax IEEE, 2021, pp.
  41--50.

\bibitem{li2022autoware_perf}
Z.~Li, A.~Hasegawa, and T.~Azumi, ``Autoware\_perf: A tracing and performance
  analysis framework for ros 2 applications,'' \emph{Journal of Systems
  Architecture}, vol. 123, p. 102341, 2022.

\bibitem{bedard2022ros2tracing}
C.~B{\'e}dard, I.~L{\"u}tkebohle, and M.~Dagenais, ``ros2\_tracing:
  Multipurpose low-overhead framework for real-time tracing of ros 2,''
  \emph{IEEE Robotics and Automation Letters}, vol.~7, no.~3, pp. 6511--6518,
  2022.

\bibitem{giraldeau2015wait}
F.~Giraldeau and M.~Dagenais, ``Wait analysis of distributed systems using
  kernel tracing,'' \emph{IEEE Transactions on Parallel and Distributed
  Systems}, vol.~27, no.~8, pp. 2450--2461, 2015.

\bibitem{maruyama2016exploring}
Y.~Maruyama, S.~Kato, and T.~Azumi, ``Exploring the performance of ros2,'' in
  \emph{Proceedings of the 13th International Conference on Embedded Software},
  2016, pp. 1--10.

\bibitem{gutierrez2018towards}
C.~S.~V. Guti{\'e}rrez, L.~U.~S. Juan, I.~Z. Ugarte, and V.~M. Vilches,
  ``Towards a distributed and real-time framework for robots: Evaluation of ros
  2.0 communications for real-time robotic applications,'' \emph{arXiv preprint
  arXiv:1809.02595}, 2018.

\bibitem{puck2020distributed}
L.~Puck, P.~Keller, T.~Schnell, C.~Plasberg, A.~Tanev, G.~Heppner,
  A.~R{\"o}nnau, and R.~Dillmann, ``Distributed and synchronized setup towards
  real-time robotic control using ros2 on linux,'' in \emph{2020 IEEE 16th
  International Conference on Automation Science and Engineering (CASE)}.\hskip
  1em plus 0.5em minus 0.4em\relax IEEE, 2020, pp. 1287--1293.

\bibitem{pardo2003omg}
G.~Pardo-Castellote, ``Omg data-distribution service: Architectural overview,''
  in \emph{23rd International Conference on Distributed Computing Systems
  Workshops, 2003. Proceedings.}\hskip 1em plus 0.5em minus 0.4em\relax IEEE,
  2003, pp. 200--206.

\bibitem{kronauer2021latency}
T.~Kronauer, J.~Pohlmann, M.~Matth{\'e}, T.~Smejkal, and G.~Fettweis, ``Latency
  analysis of ros2 multi-node systems,'' in \emph{2021 IEEE International
  Conference on Multisensor Fusion and Integration for Intelligent Systems
  (MFI)}.\hskip 1em plus 0.5em minus 0.4em\relax IEEE, 2021, pp. 1--7.

\bibitem{jiang2020message}
Z.~Jiang, Y.~Gong, J.~Zhai, Y.-P. Wang, W.~Liu, H.~Wu, and J.~Jin, ``Message
  passing optimization in robot operating system,'' \emph{International Journal
  of Parallel Programming}, vol.~48, no.~1, pp. 119--136, 2020.

\bibitem{wang2019tzc}
Y.-P. Wang, W.~Tan, X.-Q. Hu, D.~Manocha, and S.-M. Hu, ``Tzc: Efficient
  inter-process communication for robotics middleware with partial
  serialization,'' in \emph{2019 IEEE/RSJ International Conference on
  Intelligent Robots and Systems (IROS)}.\hskip 1em plus 0.5em minus
  0.4em\relax IEEE, 2019, pp. 7805--7812.

\bibitem{puck2021performance}
L.~Puck, P.~Keller, T.~Schnell, C.~Plasberg, A.~Tanev, G.~Heppner, A.~Roennau,
  and R.~Dillmann, ``Performance evaluation of real-time ros2 robotic control
  in a time-synchronized distributed network,'' in \emph{2021 IEEE 17th
  International Conference on Automation Science and Engineering (CASE)}.\hskip
  1em plus 0.5em minus 0.4em\relax IEEE, 2021, pp. 1670--1676.

\bibitem{performancetest}
\BIBentryALTinterwordspacing
Apex.AI, ``performance\_test.'' [Online]. Available:
  \url{https://gitlab.com/ApexAI/performance\_test}
\BIBentrySTDinterwordspacing

\bibitem{irobotros2performance}
\BIBentryALTinterwordspacing
iRobot, ``irobot ros 2 performance evaluation framework.'' [Online]. Available:
  \url{https://github.com/irobot-ros/ros2-performance}
\BIBentrySTDinterwordspacing

\bibitem{witte2021inferred}
T.~Witte and M.~Tichy, ``Inferred interactive controls through provenance
  tracking of ros message data,'' in \emph{2021 IEEE/ACM 3rd International
  Workshop on Robotics Software Engineering (RoSE)}.\hskip 1em plus 0.5em minus
  0.4em\relax IEEE, 2021, pp. 67--74.

\bibitem{gregg2020systems}
B.~Gregg, \emph{Systems Performance: Enterprise and the Cloud}, 2nd~ed.\hskip
  1em plus 0.5em minus 0.4em\relax Pearson, 2020.

\bibitem{ros2docsexecutor}
\BIBentryALTinterwordspacing
``Executors.'' [Online]. Available:
  \url{https://docs.ros.org/en/rolling/Concepts/About-Executors.html}
\BIBentrySTDinterwordspacing

\bibitem{peeck2021online}
J.~Peeck, J.~Schlatow, and R.~Ernst, ``Online latency monitoring of
  time-sensitive event chains in safety-critical applications,'' in \emph{2021
  Design, Automation \& Test in Europe Conference \& Exhibition (DATE)}.\hskip
  1em plus 0.5em minus 0.4em\relax IEEE, 2021, pp. 539--542.

\bibitem{casini2019response}
D.~Casini, T.~Bla{\ss}, I.~L{\"u}tkebohle, and B.~B. Brandenburg,
  ``Response-time analysis of ros 2 processing chains under reservation-based
  scheduling,'' in \emph{31st Euromicro Conference on Real-Time Systems (ECRTS
  2019)}.\hskip 1em plus 0.5em minus 0.4em\relax Schloss
  Dagstuhl-Leibniz-Zentrum fuer Informatik, 2019.

\bibitem{tang2020response}
Y.~Tang, Z.~Feng, N.~Guan, X.~Jiang, M.~Lv, Q.~Deng, and W.~Yi, ``Response time
  analysis and priority assignment of processing chains on ros2 executors,'' in
  \emph{2020 IEEE Real-Time Systems Symposium (RTSS)}.\hskip 1em plus 0.5em
  minus 0.4em\relax IEEE, 2020, pp. 231--243.

\bibitem{blass2021automatic}
T.~Blass, A.~Hamann, R.~Lange, D.~Ziegenbein, and B.~B. Brandenburg,
  ``Automatic latency management for ros 2: Benefits, challenges, and open
  problems,'' in \emph{2021 IEEE 27th Real-Time and Embedded Technology and
  Applications Symposium (RTAS)}.\hskip 1em plus 0.5em minus 0.4em\relax IEEE,
  2021, pp. 264--277.

\bibitem{blass2021ros}
T.~Bla{\ss}, D.~Casini, S.~Bozhko, and B.~B. Brandenburg, ``A ros 2
  response-time analysis exploiting starvation freedom and execution-time
  variance,'' in \emph{2021 IEEE Real-Time Systems Symposium (RTSS)}.\hskip 1em
  plus 0.5em minus 0.4em\relax IEEE, 2021, pp. 41--53.

\bibitem{yang2020exploring}
Y.~Yang and T.~Azumi, ``Exploring real-time executor on ros 2,'' in \emph{2020
  IEEE International Conference on Embedded Software and Systems
  (ICESS)}.\hskip 1em plus 0.5em minus 0.4em\relax IEEE, 2020, pp. 1--8.

\bibitem{rtwgreferencesystem}
\BIBentryALTinterwordspacing
{ROS 2 Real-Time Working Group}, ``Reference system.'' [Online]. Available:
  \url{https://github.com/ros-realtime/reference-system}
\BIBentrySTDinterwordspacing

\bibitem{kato2018autoware}
S.~Kato, S.~Tokunaga, Y.~Maruyama, S.~Maeda, M.~Hirabayashi, Y.~Kitsukawa,
  A.~Monrroy, T.~Ando, Y.~Fujii, and T.~Azumi, ``Autoware on board: Enabling
  autonomous vehicles with embedded systems,'' in \emph{2018 ACM/IEEE 9th
  International Conference on Cyber-Physical Systems (ICCPS)}.\hskip 1em plus
  0.5em minus 0.4em\relax IEEE, 2018, pp. 287--296.

\bibitem{kato2015open}
S.~Kato, E.~Takeuchi, Y.~Ishiguro, Y.~Ninomiya, K.~Takeda, and T.~Hamada, ``An
  open approach to autonomous vehicles,'' \emph{IEEE Micro}, vol.~35, no.~6,
  pp. 60--68, 2015.

\bibitem{desnoyers2006lttng}
M.~Desnoyers and M.~R. Dagenais, ``The lttng tracer: A low impact performance
  and behavior monitor for gnu/linux,'' in \emph{OLS (Ottawa Linux Symposium)},
  vol. 2006.\hskip 1em plus 0.5em minus 0.4em\relax Citeseer, 2006, pp.
  209--224.

\bibitem{gebai2018survey}
M.~Gebai and M.~R. Dagenais, ``Survey and analysis of kernel and userspace
  tracers on linux: Design, implementation, and overhead,'' \emph{ACM Computing
  Surveys (CSUR)}, vol.~51, no.~2, pp. 1--33, 2018.

\bibitem{Luetkebohle2017}
\BIBentryALTinterwordspacing
I.~Lütkebohle, ``Determinism in ros – or when things break /sometimes/ and
  how to fix it…,'' in \emph{ROSCon Vancouver 2017}.\hskip 1em plus 0.5em
  minus 0.4em\relax Open Robotics, September 2017. [Online]. Available:
  \url{https://doi.org/10.36288/ROSCon2017-900789}
\BIBentrySTDinterwordspacing

\bibitem{ros1tracetools}
\BIBentryALTinterwordspacing
{Bosch Corporate Research}, ``Ros 1 tracetools.'' [Online]. Available:
  \url{https://github.com/boschresearch/ros1\_tracetools}
\BIBentrySTDinterwordspacing

\bibitem{yang1988critical}
C.-Q. Yang and B.~P. Miller, ``Critical path analysis for the execution of
  parallel and distributed programs,'' in \emph{The 8th International
  Conference on Distributed}.\hskip 1em plus 0.5em minus 0.4em\relax IEEE
  Computer Society, 1988, pp. 366--367.

\bibitem{duda1987estimating}
A.~Duda, G.~Harrus, Y.~Haddad, and G.~Bernard, ``Estimating global time in
  distributed systems.'' in \emph{ICDCS}, vol.~87, 1987, pp. 299--306.

\bibitem{poirier2010accurate}
B.~Poirier, R.~Roy, and M.~Dagenais, ``Accurate offline synchronization of
  distributed traces using kernel-level events,'' \emph{ACM SIGOPS Operating
  Systems Review}, vol.~44, no.~3, pp. 75--87, 2010.

\bibitem{jabbarifar2014liana}
M.~Jabbarifar and M.~Dagenais, ``Liana: Live incremental time synchronization
  of traces for distributed systems analysis,'' \emph{Journal of network and
  computer applications}, vol.~45, pp. 203--214, 2014.

\bibitem{gelle2021combining}
L.~Gelle, N.~Ezzati-Jivan, and M.~R. Dagenais, ``Combining distributed and
  kernel tracing for performance analysis of cloud applications,''
  \emph{Electronics}, vol.~10, no.~21, p. 2610, 2021.

\bibitem{santos2019static}
A.~Santos, A.~Cunha, and N.~Macedo, ``Static-time extraction and analysis of
  the ros computation graph,'' in \emph{2019 Third IEEE international
  conference on robotic computing (IRC)}.\hskip 1em plus 0.5em minus
  0.4em\relax IEEE, 2019, pp. 62--69.

\bibitem{tracetoolsanalysis}
\BIBentryALTinterwordspacing
``tracetools\_analysis.'' [Online]. Available:
  \url{https://gitlab.com/ros-tracing/tracetools\_analysis}
\BIBentrySTDinterwordspacing

\bibitem{bedard2019messageflow}
\BIBentryALTinterwordspacing
C.~Bédard, ``Message flow analysis for ros through tracing,'' 2019. [Online].
  Available: \url{https://christophebedard.com/ros-tracing-message-flow/}
\BIBentrySTDinterwordspacing

\bibitem{eprosimafastdds}
\BIBentryALTinterwordspacing
eProsima, ``Fast dds.'' [Online]. Available:
  \url{https://github.com/eProsima/Fast-DDS}
\BIBentrySTDinterwordspacing

\bibitem{eclipsecyclone}
\BIBentryALTinterwordspacing
``Eclipse cyclone dds.'' [Online]. Available:
  \url{https://github.com/eclipse-cyclonedds/cyclonedds}
\BIBentrySTDinterwordspacing

\bibitem{tracecompass}
\BIBentryALTinterwordspacing
``Eclipse trace compass.'' [Online]. Available:
  \url{https://www.eclipse.org/tracecompass/}
\BIBentrySTDinterwordspacing

\bibitem{mills1991internet}
D.~L. Mills, ``Internet time synchronization: the network time protocol,''
  \emph{IEEE Transactions on communications}, vol.~39, no.~10, pp. 1482--1493,
  1991.

\bibitem{labbe2019rtab}
M.~Labb{\'e} and F.~Michaud, ``Rtab-map as an open-source lidar and visual
  simultaneous localization and mapping library for large-scale and long-term
  online operation,'' \emph{Journal of Field Robotics}, vol.~36, no.~2, pp.
  416--446, 2019.

\end{thebibliography}

\end{document}